\documentclass{ecai}
\usepackage{graphicx}
\usepackage{latexsym}

\usepackage{amsmath}
\usepackage{amsthm}
\usepackage{amssymb}
\usepackage{bbm}
\usepackage{mathtools}
\usepackage{microtype}
\usepackage{subfigure}
\usepackage{todonotes}
\usepackage{xspace}
\usepackage{booktabs}
\usepackage{hyperref}

\newtheorem{example}{Example}

\newtheorem{property}{Property}
\newtheorem{proposition}{Proposition}
\newtheorem{definition}{Definition}
\begin{document}

\begin{frontmatter}

\title{Argument Attribution Explanations in\\Quantitative Bipolar Argumentation Frameworks\\(Technical Report)}

\author[A]{\fnms{Xiang}~\snm{Yin}\orcid{0000-0002-6096-9943}\thanks{Corresponding Author. Email: x.yin20@imperial.ac.uk}}
\author[B,A]{\fnms{Nico}~\snm{Potyka}\orcid{0000-0003-1749-5233}}
\author[A]{\fnms{Francesca}~\snm{Toni}\orcid{0000-0001-8194-1459}}

\address[A]{Imperial College London, UK}
\address[B]{Cardiff University, UK}

\begin{abstract}
Argumentative explainable AI has been advocated by several in recent years, with an increasing interest on explaining the reasoning outcomes of Argumentation Frameworks (AFs).
While there is a considerable body of research on qualitatively explaining the reasoning outcomes of AFs with debates/disputes/dialogues 
in the spirit of %under IS NOT CORRECT, AS ALSO GRADUAL SEMANTICS USED IN DEBATES-LIKE EXPLANATIONS 
\emph{extension-based semantics}, explaining the quantitative reasoning outcomes of AFs under \emph{gradual semantics} has not received much attention, despite widespread use in applications.
In this paper, we contribute to filling this gap by proposing a novel theory of \emph{Argument Attribution Explanations (AAEs)} by incorporating the spirit of feature attribution from machine learning in the context of Quantitative Bipolar Argumentation Frameworks (QBAFs): whereas feature attribution is used to determine the influence of features towards outputs of machine learning models,
AAEs are used to determine the influence of arguments towards \emph{topic argument}s of interest.
We study desirable properties of AAEs, including some new ones and some partially adapted from the literature to our setting.
To demonstrate the applicability of our AAEs in practice,
we conclude by carrying out two case studies in the scenarios of fake news detection and movie recommender systems.
\end{abstract}
\end{frontmatter}

\section{Introduction}
\label{sec_intro}
Explainable AI (XAI) is playing an increasingly important role in AI
towards %its 
 safety, reliability and trustworthiness \cite{adadi2018peeking}. 
% \todo{Would recommend "various methods" instead of "two methods" because there are definitely more than two and people may be offended if their method is ignored}
%A variety of 
Various methods %are popular 
have been proposed in this field for providing explanations for several AI algorithms, models, and systems (e.g. see recent overviews \cite{minh2022explainable,adadi2018peeking}).

A %big 
popular category of explanation methods is \emph{feature attribution}, %which is basically
aiming at assigning a ``feature importance score" to each input feature fed to the AI of interest (notably machine learning models), denoting %the contribution of the input 
its contribution 
to the output decision by the AI. %Some famous 
Feature attribution methods include, amongst others, LIME \cite{ribeiro2016should}, SHAP \cite{lundberg2017unified}, SILO \cite{bloniarz2016supervised} and gradient-scoring \cite{baehrens2010explain}.
%\todo{add references after the names}.
Explanations returned by feature attribution methods are intuitive in %the sense that the explanation of the output is just attributed to the input features with feature importance scores, which is
that they focus on explaining the outputs in terms of the inputs alone, making it unnecessary to go into the details of the inner mechanism of the %machine learning models
underlying AI. 
Furthermore, 
%What is more, it is 
feature attribution explanations are easy for people to understand by just checking the positive or negative influence of the input features towards the outputs and the ranking of the magnitude of the scores.

%Another category of explanation methods in the literature is
Alongside feature attribution  methods,
in recent years \emph{argumentative XAI} %has gradually shown 
is increasingly showing 
%its advantages %in the explanation field 
benefits
for various forms of AI  (see  \cite{vcyras2021argumentative,vassiliades2021argumentation} for overviews).
Basically, argumentative XAI applies computational argumentation~\cite{arg-survey}  %theory
to extract  \emph{argumentation frameworks (AFs)} as  skeletons for %explainer for machine learning algorithms or systems
explanations% for various forms of AI
. For example, \cite{vcyras2019argumentation}
uses abstract AFs~\cite{Dung_95} to explain the outputs of schedulers, 
\cite{potyka2021interpreting} proposes to use weighted bipolar AFs to explain %the random forest algorithm 
multi-layer perceptrons and \cite{cocarascu2019extracting} propose to use  \emph{Quantitative Bipolar AFs (QBAFs)}~\cite{baroni2015automatic} to explain %the recommender systems 
movie review aggregations.
%
%For argumentative explanation, the argumentation framework itself is the explanation.
%AFs can be seen as explanations, which are intuitive in the light of showing 
Whereas feature attribution methods
focus on the input-output behaviour of the underlying AI,
AFs as explanations point to the 
dialectical relationships among %inner 
arguments, abstractly representing interactions among the inner components of the underlying AI %with the AF encoding the underlying %machine learning model
encoded by the AFs. %Based on the natural mechanism of the AF, it is easy 
These AFs provide a natural mechanism for users to interact with %the users and 
the AI \cite{rago2020argumentation} and
may help find irrationalities in the underlying AI to aid debugging and improving the AI \cite{garcia2007dialectical}.%\todo{add a reference here? The only that comes to my mind is Piyawat's FIND, but he does not use argumentation explicitly, anybody else?}

% %Although these two 
% These two categories of
% explanation methods have %a number of 
% complementary strengths and weaknesses%advantages% and show huge success in XAI field, some problems still need to be further considered
% .  
% Feature attribution methods give intuitive but``flat" explanations, because, except for the feature attribution scores, they do not provide information about the inner mechanism of the underlying AI, %which is still vague 
% and thus cannot help much for %the model 
% interacting with and debugging %and improving
% the AI.

\iffalse REWRITTEN BELOW TO BE IN THE SPIRIT OF THE ABSTRACT - NOT DISPUTES BUT...
Although argumentative XAI %is still vague to some extent because it only 
provides information about the AI inner mechanism in the form of dialectical relationships among arguments and the AF semantics, 
\fi

Existing forms of argumentative XAI
 are predominantly \emph{qualitative} in that they focus on
explaining
the reasoning outcomes of AFs with debates/disputes/dialogues in
the spirit of \emph{extension-based semantics} (e.g. as in \cite{Dung_95}). These qualitative explanations mirror 
interactions within the inner mechanism of the underpinning AI as dialectical exchanges between arguments. For example,  \cite{vcyras2019argumentation} use `explanation via (non-)attacks' and \cite{cocarascu2019extracting} define explanations as template-driven dialogues using attacks and supports in the AFs. 
Instead, 
explaining the \emph{quantitative}
reasoning outcomes of AFs under \emph{gradual semantics} (e.g. those proposed in \cite{baroni2015automatic,potyka2021interpreting}) has not received
much attention, in spite of the widespread use of this form of semantics in several applications (e.g. fake news detection~\cite{kotonya2019gradual}, movie recommendations~\cite{cocarascu2019extracting} and fraud detection~\cite{chi2021optimized}). However, 
in many application settings, it is important to see how arguments in AFs topically influence one another, and %without
how much positive/negative influence is transmitted from one argument to another.  
 This is especially the case when explanations are needed for a \emph{topic argument} of interest (e.g. an argument corresponding to the output of a %machine learning 
 classifier as in \cite{albini2020deep,potyka2021interpreting}) and it is essential to assess %Thus, it is unclear 
which arguments have more importance towards the topic argument.

\iffalse ALREADY CAPTURED ABOVE 
%Therefore, 
In other words, 
although AFs %are capable of providing 
naturally provide \emph{qualitative} dialectical explanations, 
it is still worth exploring
whether they can provide \emph{quantitative} explanations% of AFs to further boost their explainability
.
\fi

\begin{figure}[ht]
	\centering
		\includegraphics[width=0.8\columnwidth]{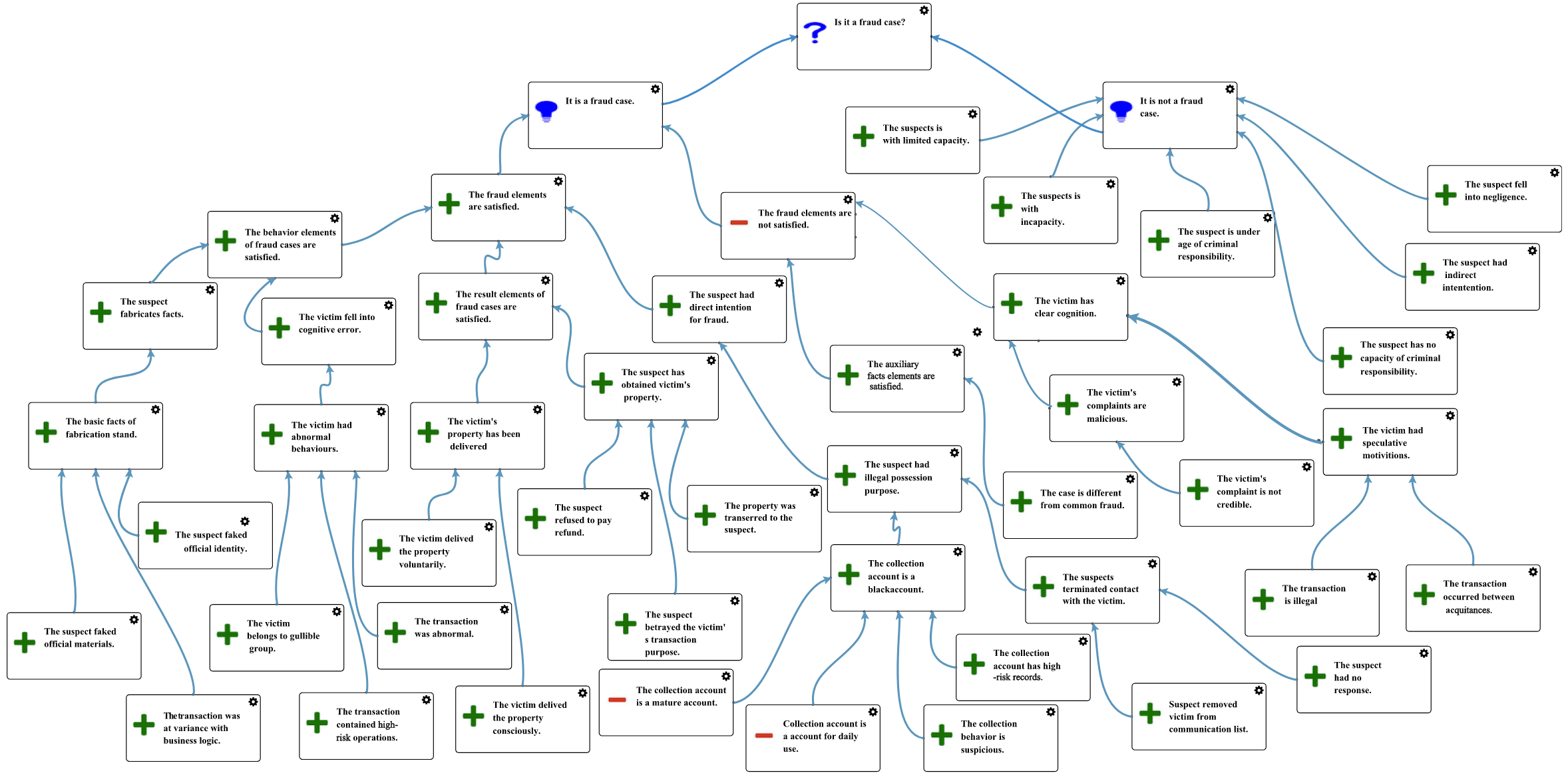}
	\caption{Fraud detection in e-commerce%transactions
 ~\cite{chi2021optimized}.
(Our emphasis is 
on the QBAF's size/complexity rather than contents, so readability is not a concern%We are intend to show the complicacy of the QBAF rather than its contents
).}
	\label{fig_big_AF}
\end{figure}

In this paper, we contribute to filling this gap by proposing a novel theory of \emph{Argument Attribution Explanations (AAEs)} by incorporating the spirit of feature attribution from machine learning in the context of QBAFs.
% Motivated by the merits of feature attribution explanations, we believe argumentative explanations can be made more powerful by incorporating the idea of feature attribution methods. % to computational argumentation, which naturally combines the complementary strength of these two explanation methods
% by combining their. 
% Therefore, in this paper, we propose a theory to compute the attribution between arguments based on the existing dialectical relations in the extracted AF.
% In this paper, we formally propose the theory of argument attribution explanation, a powerful explanation method, by applying the spirit of the feature attribution methods to computational argumentation.
% \todo{I would reconsider the previous sentence.
% It seems more accurate to write that we are applying
% the idea of feature attribution methods to 
% argumentation. One could argue that this could be
% used for argumentative XAI, but since this is not
% really part of this paper, I would move a discussion
% of this topic either to "future work" or to the
% initial motivation}
%First, 
With respect to %dialectical 
qualitative explanations alone, 
AAEs %are more intuitive in that they are able to quantitatively evaluate 
allow to measure and compare the contribution of different arguments towards topic arguments in QBAFs under the Discontinuity Free Quantitative Argumentation Debate (\emph{DF-QuAD}) gradual semantics~\cite{rago2016discontinuity}, even when the comparison is difficult with qualitative explanations alone.
%Especially in some settings with large AFs, dialectical explanations alone may be lengthy. 
This is the case with large QBAFs as visualized in Figure~\ref{fig_big_AF}  %shows applying a QBAF to solve the problem of 
%for fraud detection in e-commerce transactions
from~\cite{chi2021optimized}, %the dialogical explanations from leaf arguments to the root argument are somewhat cumbersome and lack of intuitiveness.
where it is hard to see how quantitative explanations can be %cumbersome and unintuitive.
intuitively delivered for the children of the root as topic arguments.\footnote{AAEs for this example can be found 
in 
% \url{https://arxiv.org/abs/2307.13582}.}
the Supplementary Material (SM).} %%% ARXIV VERSION
Additionally, AAEs %are more suitable for explaining quantitative AFs than non-quantitative ones because they 
take the \emph{base scores} of arguments in QBAFs into account% when explaining
. Different base scores should give rise to different explanations, but %dialogical 
qualitative explanations %are always identical no matter what the 
disregard base scores, as they only consider the  QBAFs' structure regardless of quantitative information%, which is an obvious limitation for explaining quantitative AFs
.

In this paper, we %first 
formalize AAEs and %define four different types of connectivity in QBAF (i.e. disconnectivity, direct, indirect and multifold connectivity). Then, we 
analyze 
% the mathematical properties of AAEs under four different types of connectivity in QBAF (i.e. disconnectivity, direct, indirect and multifold connectivity).
some qualitative and quantitative guarantees for different types of \emph{connectivity} in QBAFs.
%direct and indirect connectivity from both qualitative and quantitative perspectives
We then study several desirable properties of AAEs as explanations%from the \emph{faithfulness} WHERE? and \emph{computational complexity} aspects
, including some adapted from the literature and some novel ones.
Finally, we show the applicability of AAEs in two scenarios:  fake news detection~\cite{kotonya2019gradual} and movie recommender systems~\cite{cocarascu2019extracting}.
Overall, 
the contribution of this paper is threefold.
\begin{itemize}
    \item We propose the novel theory of %argument attribution explanation 
    AAEs (Section \ref{sec_theory}). %by incorporating the idea of feature attribution method;
    \item We study (new and adapted) desirable properties of AAEs as explanations (Section \ref{sec_properties}).
    \item We show applicability of AAEs in practice (Section \ref{sec_cases}).
\end{itemize}
% (1) gradient is computationally efficient;\\
% (2) can help to build compact AF by remove unimportant arguments.\\
% (3) faithful?\\
% (4) intuitive + interactive
%
% The rest of this paper is structured as follows. 
% In section 2, we review the definition about QBAF and DF-QuAD gradual semantics.
% In section 3, we give the formal definition about our theory.
% In section 4, we analyse the property of this method.
% In section 5, we give 2 case studies to show the real-world application from the practical perspective.
% Finally, we conclude in the last section.
%%%%%%%%
The proofs of all results %in the paper can be found 
are
in 
% \url{https://arxiv.org/abs/2307.13582}.
the SM.  %%% ARXIV VERSION

\section{Related Work}
\label{sec_related}
\paragraph{Argumentative XAI} 
Following~\cite{vcyras2021argumentative},  we can distinguish two types of argumentative explanations.
One is %\textbf
{\emph{intrinsic}} argumentative explanations, whereby the explained underlying models themselves are already AFs. 
% \cite{cocarascu2020data,vcyras2019explanations,kokciyan2020argumentation}
% \todo{add references for intrinsic}
%A case in point is some 
Examples include recommender systems  \cite{briguez2014argument,rodriguez2017educational}
built with suitable AFs specified in DeLP \cite{garcia2013formalizing}  and generating recommendations by the reasoning outcomes of the AFs. 
% \todo{add reference}
The other is %\textbf
{\emph{post-hoc}} argumentative explanations, %which focus on the 
for underlying models that are not based on AFs %or are
and are not argumentative in spirit. %Therefore, 
In order to extract argumentative explanations from these models, it is %important 
crucial to first extract AFs%from the underlying models
, that is, extract arguments and dialectical relations while identifying an appropriate argumentation semantics matching the models' behaviour. %According to 
Depending on
the representational extent of the AFs for the underlying models, post-hoc explanations can be further divided into two sub-types: %\textbf
{\emph{complete}} and %\textbf
{\emph{approximate}} argumentative explanations \cite{vcyras2021argumentative}. 
Complete argumentative explanations suit many settings, including decision-making systems~\cite{zeng2018context}, knowledge-based systems \cite{arioua2015query} and scheduling~\cite{vcyras2019argumentation}. %For example, AFs can be used to completely represent the decision-making models and explanations can be extracted from the encoded AFs \cite{zeng2018context}. 
% \todo{add reference for complete}
In the case of approximate argumentative explanations% mean 
, the mapping process from the underlying models to AFs is incomplete%~\cite{vcyras2021argumentative}
, in the sense that AFs are extracted from parts of the underlying models rather than the whole models. For example, \cite{sendi2019new} first extract rules from trained neural networks, and construct AFs based on these rules, and \cite{albini2020deep} abstract away trained neural networks as QBAFs by treating groups of neurons as arguments and understanding feature attribution methods as a gradual semantics.
%
%ADDED A COMMENT ABOUT HOW ALL THIS RELATES TO US - POSITIONING OUR WORK IN RELATED WORK
Our argumentative explanations, in the form of AAEs, assume an AF as a starting point, in the form of a QBAF under the DF-QuAD gradual semantics, and are usable alongside any existing form of argumentative XAI based on the same AFs and semantics.

% From the \emph{form} perspective of argumentative explanations, a significant principle is that explanations of the models should be easily understood by humans \cite{vcyras2021argumentative,lage2019human}. Therefore, the forms of most AF-based explanations are friendly to humans, like visual, and dialogue forms.
% In addition, based on the structure of the AFs, a number of forms of the explanations are directly extracted from the AF structure. Due to the skeleton of the AFs being built naturally from the human perspective, therefore, it is reasonable and legitimate to use the structure of AFs as the explanations. 
% For instance, dispute trees \cite{dung2007computing} are intuitive explanations extracted from AFs with strong theoretical guarantees for many properties of the explanations. Another form based on AF structure is dialogue games \cite{carlson2012dialogue}, which can also be extracted from the dispute trees.
\paragraph{Feature Attribution Methods} The intuition of feature attribution methods is to measure the contribution of the features to the output by feature attribution scores.
\textbf{\emph{LIME}}~\cite{ribeiro2016should} is a %locally 
model-agnostic explanation method, which can locally explain any instance (input-output pair) by  linear approximation, learning a linear surrogate model based on  sampling around the instance of interest. \textbf{\emph{SILO}} \cite{bloniarz2016supervised} shares the same idea of LIME, but %the difference lies 
differs from it in two aspects. First, LIME uses synthetic data generated by the approximation method to fit the linear surrogate model, while SILO directly uses the data from the training set. Second, for LIME, the weight of each synthetic data point is decided by the distance, while for SILO, the weight is decided by calling a random forest classifier. 
% \textbf{\emph{MAPLE}} is an improvement of SILO. It basically first calls SILO algorithm and then applies DSTump \cite{kazemitabar2017variable} to choose the important features. 
% In this way, MAPLE can not only server as an explanation algorithm but also a prediction model. 
\textbf{\emph{SHAP}}~\cite{lundberg2017unified} is another popular model-agnostic feature attribution method% in recent years
, which is theoretically guaranteed by game theory \cite{fudenberg1991game} and satisfies several desirable properties. SHAP computes the marginal contribution of the features as their attribution scores. However, SHAP is computationally inefficient because of the combinatorial 
%case of different 
combination of
features. \textbf{\emph{Gradient-scoring}}~\cite{baehrens2010explain}
%\todo{we shouldn't cite our own paper here, gradient scoring has been introduced in 2010 already (we just proposed a formal justification of the approach), see the "baehrens2010explain" reference in our gradient paper} 
is also an attribution method especially suitable for differentiable classifiers with continuous data, which demonstrates a greater accuracy and efficiency than LIME, SHAP and other popular attribution methods in such settings \cite{potyka2022towards}.
%
%ADDED A COMMENT ABOUT HOW ALL THIS RELATES TO US - POSITIONING OUR WORK IN RELATED WORK
Our AAEs are in the spirit of gradient scoring, but take a QBAF under the DF-QuAD semantics as a starting point.

\paragraph{Properties of Explanations} \cite{chen2022makes}
%\todo{year seems to be missing in the chen reference} 
summarize %a majority of 
popular properties of explanations in the literature into four categories.
Two of them are related to this paper.
The first category is %\textbf
{\emph{robustness/sensitivity}}, %which guarantees 
amounting to
the robustness of explanation methods under small perturbations of the inputs. 
\iffalse 
\cite{potyka2022towards} %propose that for 
indicates that gradient-based attribution methods %, they 
may violate %the robustness 
this property %in the 
with non-linear underlying models because they can only capture local behavior of the models.\todo{why is the previous sentence interesting/relevant?} 
\fi
The second %property
category is %\textbf
{\emph{faithfulness/fidelity}}, %measuring 
concerning the loss between the explanation model and the underlying model. Many properties in the literature are proposed to measure faithfulness, which is %especially important 
core for attribution explanation methods \cite{chen2022makes}. 

For argumentative explanations, with a few exceptions (notably~\cite{amgoud2022axiomatic}), properties are not %very 
well-studied so far \cite{vcyras2021argumentative}. Some of the existing properties borrow ideas behind general explanation properties. For instance, \emph{fidelity} \cite{vcyras2019argumentation} is used to measure whether the extracted AF is faithful to the underlying model% or system
. Also,  \emph{transparency} and \emph{trust} are studied as cognition-related properties of explanations% which are mentioned above
%by some
~\cite{rago2020argumentation}.
%
%ADDED A COMMENT ABOUT HOW ALL THIS RELATES TO US - POSITIONING OUR WORK IN RELATED WORK
We contribute some novel properties of argumentative explanations while also adapting some existing ones for argumentative explanations (e.g. \emph{explainability} from ~\cite{amgoud2022axiomatic}) and %some 
for standard feature attribution (e.g. faithfulness). 

\section{Background}
\label{sec_background}
We recall the definition of QBAF \cite{baroni2015automatic}, a form of quantitative bipolar AFs~\cite{amgoud2008bipolarity}, and the DF-QuAD gradual semantics~\cite{rago2016discontinuity}.

\begin{figure}[t]
	\centering
		\includegraphics[width=0.6\columnwidth]{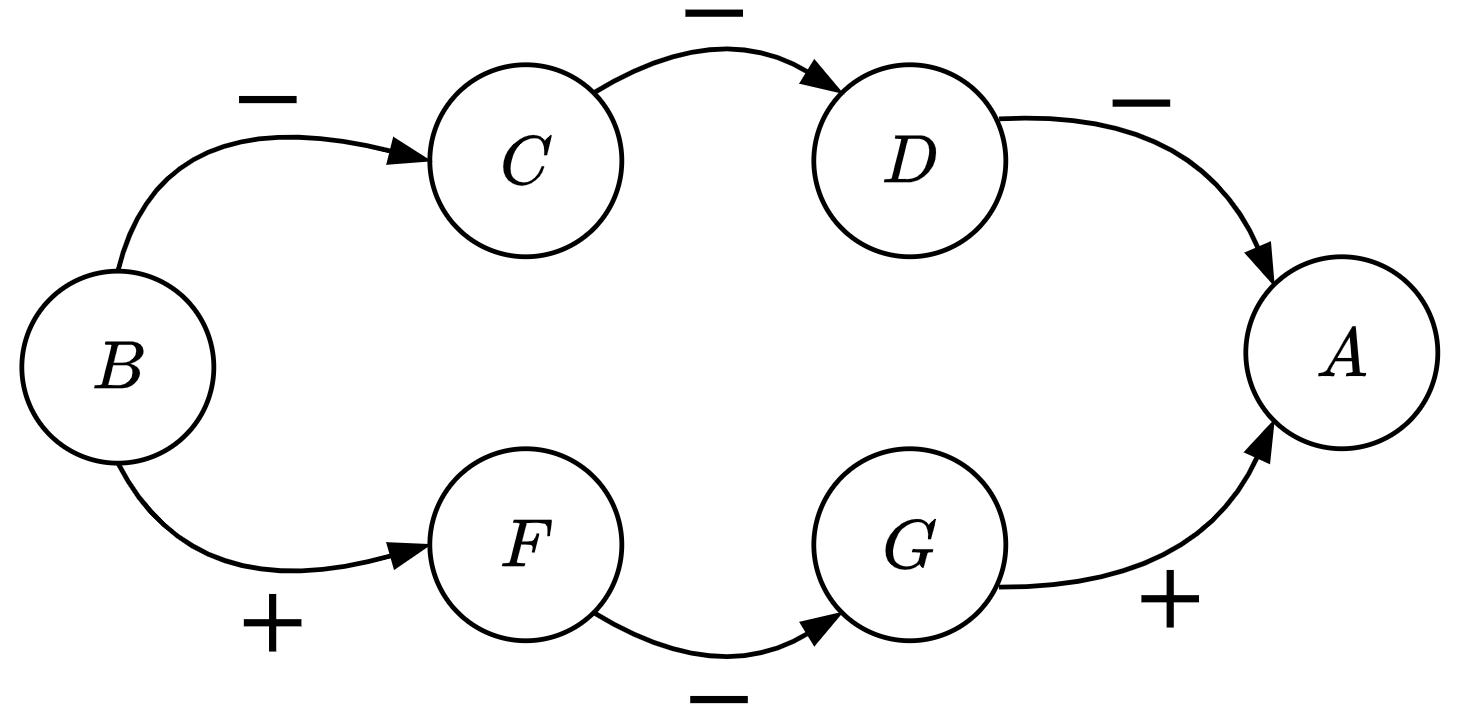}
	\caption{An example QBAF structure.}
	\label{fig_bcda_two_path}
\end{figure}

\begin{definition}[QBAF]
\label{def_QBAF}
A \emph{QBAF} is a quadruple $\mathcal{Q}=\left\langle\mathcal{A}, \mathcal{R}^{-}, \mathcal{R}^{+}, \tau \right\rangle$, where $\mathcal{A}$ is the set of \emph{arguments}; $\mathcal{R}^{-}$ is the \emph{attack} relation ($\mathcal{R}^{-} \subseteq \mathcal{A} \times \mathcal{A} $); $\mathcal{R}^{+}$ is the  \emph{support} relation ($\mathcal{R}^{+} \subseteq \mathcal{A} \times \mathcal{A} $); $\mathcal{R}^{-}$ and $\mathcal{R}^{+}$ are disjoint; $\tau$ is the \emph{base score} %function 
($\tau\!:\! \mathcal{A} \!\rightarrow \! [0,1]$).
\end{definition}
%In Definition \ref{def_QBAF}, 
%Here, $\tau(A)$ is the base score for argument $A \in \mathcal{A}$.
We often denote the structure of QBAFs graphically, where nodes
represent the arguments and edges the support and 
attack relations. We label edges in attack and support relations with $-$ and
$+$, respectively. Figure \ref{fig_bcda_two_path} shows an example.
A QBAF is called \emph{acyclic} if the corresponding graph is acyclic.
As in \cite{rago2016discontinuity}, we restrict attention to \emph{acyclic QBAFs}. 

The \emph{dialectical strength} of arguments in QBAFs can be evaluated by several gradual semantics 
$\sigma: \mathcal{A} \rightarrow [0,1]$, 
e.g. as defined in \cite{baroni2015automatic,amgoud2018evaluation,Potyka18,potyka2021interpreting}. 
%\todo{is this a good way to better motivate DF-QuAD?}
The explainability of some gradual semantics like the h-categorizer semantics~\cite{besnard2001logic} and the counting semantics~\cite{pu2015attacker} has been studied in~\cite{delobelle2019interpretability}.
In this paper, we focus on %the challenging problem of 
explaining the %Discontinuity Free Quantitative Argumentation Debate (DF-QuAD) 
DF-QuAD gradual semantics \cite{rago2016discontinuity}. This problem has been neglected in the literature so far even though  %because it is easy to implement and deploy and 
DF-QuAD has broad applicability~\cite{rago2016discontinuity,kotonya2019gradual,cocarascu2019extracting,chi2021optimized,chi2022quantitative} in settings where explainability is important.
%
%We let $\sigma(A)$ denotes the dialectical strength of argument $A \in \mathcal{A}$, where strength function $\sigma: \mathcal{A} \rightarrow \mathbb{I}$ is referred to as a gradual semantics.
% \todo{separate this sentence into two sentences: the first one explains tau (which is defined in the definition). The second one explains sigma (which is not defined in the definition)}
%Typically, $\mathbf{QBAF}$ \cite{baroni2019fine} is an extension of $\mathbf{BAF}$ \cite{amgoud2008bipolarity} with base scores.
% \todo{add a reference for BAFs, e.g.,
% Leila Amgoud, Claudette Cayrol, Marie-Christine Lagasquie-Schiex, P. Livet:
% On bipolarity in argumentation frameworks. Int. J. Intell. Syst. 23(10): 1062-1093 (2008)}
%QBAF assigns base scores to arguments and then apply semantics to compute the dialectical strength of each arguments.
%In the literature, three gradual semantics are popular. They are Quantitative Argumentation Debate (QuAD) \cite{baroni2015automatic}, Discontinuity Free Quantitative Argumentation Debate (DF-QuAD) \cite{rago2016discontinuity}, and the Restricted Euler-based (REB) \cite{amgoud2018evaluation} semantics. Here, we will only introduce the DF-QuAD which we use 
%is the most related in the setting of  in this paper.
%

In DF-QuAD, for any argument $A$, %$\tau(A)$ and $\sigma(A)$ are its base score and strength, respectively. 
$\sigma(A)$ is defined %by strength function 
as follows:
$$
\sigma(A)= 
    \begin{cases}
        \tau(A)-\tau(A)\cdot(v_{Aa}-v_{As}) & if\ v_{Aa} \geq v_{As}\\
        \tau(A)+(1-\tau(A))\cdot(v_{As}-v_{Aa}) & if\ v_{Aa} < v_{As}\\
    \end{cases}
$$
where $v_{Aa}$ is the \emph{aggregation strength} of all the attackers against $A$, while $v_{As}$ is the \emph{aggregation strength} of all the supporters for $A$.
% Suppose $n$ attackers $X_{n}$ are against $A$ with strength $\sigma(X_{n})$; $m$ supporters $Z_{m}$ are for $A$ with strength $\sigma(Z_{m})$. 
$v_{Aa}$ and $v_{As}$ are defined %by aggregation 
as follows:
$$
v_{Aa}=1-\prod_{\left \{ X \in \mathcal{A} \mid (X,A) \in \mathcal{R^{-}} \right \} }(1-\sigma(X));
$$
$$
v_{As}=1-\prod_{\left \{ X \in \mathcal{A} \mid (X,A) \in \mathcal{R^{+}} \right \} }(1-\sigma(X)).
$$
An example of applying DF-QuAD is %shown in Example \ref{example_DF_QuAD}
as follows.
\begin{example}
\label{example_DF_QuAD}
Consider the acyclic QBAF $\mathcal{Q}=\left\langle\mathcal{A}, \mathcal{R}^{-}, \mathcal{R}^{+}, \tau \right\rangle$ % equipped with DF-QuAD gradual semantics $\sigma$ 
with arguments and relations 
as in Figure \ref{fig_bcda_two_path} and
%Suppose the base score $\tau$ of every argument is 0.5 in $\mathcal{Q}$.
$\tau(X)=0.5$ for all $X \!\in \!\mathcal{A}$.
It is easy to see that 
%According to aggregation function,
$v_{Aa}=0.375$, $v_{As}=0.125$, $v_{Ba}=v_{Bs}=0$, $v_{Ca}=0.5$, $v_{Cs}=0$, $v_{Da}=0.25$, $v_{Ds}=0$, $v_{Fa}=0$, $v_{Fs}=0.5$, $v_{Ga}=0.75$, $v_{Gs}=0$.
%According to strength function,
Then, $\sigma(A)=0.375$, $\sigma(B)=0.5$, $\sigma(C)=0.25$, $\sigma(D)=0.375$, $\sigma(F)=0.75$, $\sigma(G)=0.125.$
\end{example}
In the remainder, unless specified otherwise, we will assume %as given 
an acyclic QBAF $\mathcal{Q}\!=\!\left\langle\mathcal{A}, \mathcal{R}^{-}, \mathcal{R}^{+}, \tau \right\rangle$ and $\sigma$ given by DF-QuAD.\footnote{Note that acyclic QBAFs are not restricted to trees, as shown in Figure~\ref{fig_bcda_two_path}.
%some of the properties hold for tree-like QBAFs, they still hold in the general case.
}

\section{Argument Attribution in QBAFs}
\label{sec_theory}
We formally propose the theory of AAEs in acyclic QBAFs evaluated by DF-QuAD gradual semantics. 
Then, we define \emph{connectivity} in QBAFs and study the properties of \emph{direct} and \emph{indirect} connectivity \emph{qualitatively} and \emph{quantitatively}, which will play a role in studying the properties of AAEs as explanations in Section~\ref{sec_properties}.
% In this paper, we mainly focus on the acyclic QBAF because in real explanation applications, acyclic argumentation framework is typically able to reach the goal.
%%%%%%%
% \begin{definition}[Acceptance]
% \label{def_acceptance}
% In QBAF, if the strength of the argument is in [0, 0.5), then we call this argument is \textit{rejected};
% if the strength of the argument is in [0.5, 1], then we call this argument is \textit{accepted}.
% \end{definition}

Our AAEs are inspired %in spirit 
by gradient-based feature attribution explanations in machine learning (e.g. see \cite{shrikumar2016not,shrikumar2017learning}) and gradient-based \emph{contribution function} in \cite{vcyras2022dispute}, which, in our setting, capture the \emph{sensitivity} of the dialectical strength of a \emph{topic argument} with respect to the base score of other arguments, as follows. 
%as its attribution score.
\begin{definition}[Argument Attribution Explanations (AAEs)]
\label{def_attribution}
Let 
$A, B \in \mathcal{A}$, where $A$ is called the \emph{topic argument}.
For a perturbation $\varepsilon \in \left[-\tau(B),0) \cup (0,1-\tau(B)\right]$, 
% \todo{the interval should be $[-\tau(B), 1-\tau(B)] \setminus \{0\}$ if 
% we do not restrict to positive changes (see below)}
let $\mathcal{Q}_{\varepsilon}$ be the QBAF resulting from $\mathcal{Q}$
% by changing the base score $\tau(B)$ of $B$ to $\tau_{\varepsilon}(B) = \tau(B) + \varepsilon$, where $\varepsilon \to 0$.
by perturbing the base score $\tau(B)$ to $\tau_{\varepsilon}(B) = \tau(B) + \varepsilon$.
Let $\sigma, \sigma_{\varepsilon}$ denote %the gradual semantics corresponding to 
DF-QuAD for 
$\mathcal{Q}, \mathcal{Q}_{\varepsilon}$, respectively.
The %argument attribution 
\emph{AAE}\footnote{The well-definedness of AAE is shown in Proposition~\ref{proposition_explainability}.} from $B$ to $A$ is
% \todo{NP: why do you restrict to $0^+$ (positive changes)? This corresponds to a one-sided derivative.
% When the function is differentiable, the one-sided
% derivatives are equal to the derivative in the
% interior of the function domain, but not necessarily
% on the boundary. In your case, for a base score of $1$,
% there is no positive $\varepsilon$. You could
% set $\varepsilon = 0$, but then the influence is always $0$, which is not intended. I would just replace
% $0^+$ with $0$}
$$
\left. \nabla \right|_{B \mapsto A}=\lim_{\varepsilon \to 0}\frac{\sigma_{\varepsilon}(A)-\sigma(A)}{\varepsilon}.
$$
\end{definition}
As an illustration, consider $\mathcal{Q}$ in Figure \ref{fig_bcda_two_path}. The perturbation of $\tau(B)$ will give rise to a perturbation of $\sigma(A)$. Then, we can use the limit of the ratio of the perturbation, as defined above, to represent the %attribution score
AAE from $B$ to $A$ (see 
% \url{https://arxiv.org/abs/2307.13582} 
the SM  %%% ARXIV VERSION
for details). 
%\todo{what are they? should we not give the full example/calculation? if no space here in the appendix?}\todo{Xiang: Example 5 is added in the Appendix.}

We next distinguish three possible %influences
types of influence.
%\todo{I found this definition is the same as definition 9 in delobelle2019interpretability, do we need to mention this?}
\begin{definition}[Attribution Influence]
\label{def_influence_pos_neg}
We say that the \emph{attribution influence} from
$B$ to $A$ is
\emph{positive} if $\left. \nabla \right|_{B \mapsto A}>0$,
\emph{negative} if $\left. \nabla \right|_{B \mapsto A}<0$ and
\emph{neutral} if $\left. \nabla \right|_{B \mapsto A}=0$.
% \todo{NP: tried to simplify the definition a little bit
% (could also just be inline now). If you don't like it,
% you can just delete it and uncomment the old definition below}
%old definition
% Let $\mathcal{Q}=\left\langle\mathcal{A}, \mathcal{R}^{-}, \mathcal{R}^{+}, \tau \right\rangle$ be an acyclic QBAF, and $A, B \in \mathcal{A}$. The attribution influence between any two arguments are defined as follows:
% \begin{itemize}
%     \item If $\left. \nabla \right|_{B \mapsto A}>0$, then we call $B$ has positive influence on $A$;
%     \item If $\left. \nabla \right|_{B \mapsto A}<0$, then we call $B$ has negative influence on $A$;
%     \item If $\left. \nabla \right|_{B \mapsto A}=0$, then we call $B$ has neutral influence on $A$.
% \end{itemize}
%\todo{instead of "<equation> denotes..." better "if <equation>, then we call/say that..."}
\end{definition}

In order to analyze the influence of an argument on
a topic argument, we have to consider the different
paths connecting them.
For this purpose, we define some terminology % in the following definition
next.
\begin{definition}[Path and Path Set]
\label{def_path_and_path_set}
Let 
%$X_1, \dots,  X_n 
$X,Y \in \mathcal{A}$ and $\mathcal{R} \!=\! \mathcal{R}^{-}\! \cup \mathcal{R}^{+}$. A \emph{path} $\phi $ between %$X_1$ and $X_n$ 
$X$ and $Y$
is a sequence $ X_1, \dots, X_n$ of arguments in $\mathcal{A}$ such that  $n \geq 2$, $X_1=X, X_n=Y$ and
%$A_1$ to $A_k$ if 
$(X_i, X_{i+1}) \in \mathcal{R}$ for $1 \leq i \leq %k
n-1$.
We refer to $X_2, \dots, X_{n-1}$ as the \emph{middle arguments} and to $m_{\phi}=n-2$ as the number of middle arguments in path $\phi$.
% The length of the path $len(\phi) = k-1$ is the number of edges 
% in the path.
% For $X_{1},X_{n} \in \mathcal{A}$, 
% we let $\Phi_{B \mapsto A}$ be the set of all paths from $B$ to $A$. 
We let $\Phi_{X%_{1}
\mapsto %X_{n}
Y}$ denote the set of all paths $\phi$ from %$X_{1}$ to $X_{n}$
$X$ to $Y$, and $\left| \Phi_{X%_{1}
\mapsto %X_{n}
Y} \right|$ denote the number of paths in $\Phi_{X%_{1} 
\mapsto %X_{n}
Y}$.
% $$
% \Phi_{A_{1} \to A_{k}}=\bigcup \phi
% $$
\end{definition}

We distinguish four types of connectivity based
on the number of paths between two arguments.
\begin{definition}[Connectivity]
\label{def_connectivity}
% Let $\mathcal{Q}=\left\langle\mathcal{A}, \mathcal{R}^{-}, \mathcal{R}^{+}, \tau \right\rangle$ be an acyclic QBAF, and 
% Let $A, B \in \mathcal{A}$. 
%We define four types of connectivity from $B$ to $A$ as follows:
For any
$A, B \in \mathcal{A}$:
\begin{itemize}
    \item $B$ is \emph{disconnected} %to 
    from $A$ iff $\left| \Phi_{B \mapsto A} \right|=0$;
    \item $B$ is \emph{directly} connected to $A$ iff $\left| \Phi_{B \mapsto A} \right|=1$ and $m_{\phi}=0$ for $\phi \in \Phi_{B \mapsto A}$;
    \item $B$ is \emph{indirectly} connected to $A$ iff $\left| \Phi_{B \mapsto A} \right|=1$ and $m_{\phi} \geq 1$ for $\phi \in \Phi_{B \mapsto A}$;
    \item $B$ is \emph{multifold} connected to $A$ iff $\left| \Phi_{B \mapsto A} \right|>1$.
\end{itemize}
% \todo{maybe the terminology could be made more 
% intuitive: perhaps "directly connected" instead of "point-wise connected",
% "indirectly connected" instead of "path-wise connected" and 
% "multifold connected" instead of "graph-wise connected"?}
\end{definition}
% For illustration, Figure \ref{fig_bcda_two_path} shows an intuitive view about different connectivity types in Definition \ref{def_connectivity}. 
Here, we show an example to explain  connectivity in QBAFs. 
\begin{example}
\label{example_connectivity}
In Figure \ref{fig_bcda_two_path}, 
$C$ is disconnected %to 
from $F$;
$D$ is directly connected to $A$ because there is only one path from $D$ to $A$, and no middle arguments in between.
$C$ is indirectly connected to $A$ because there is only one single path connecting $C$ and $A$, and $D$ is the only middle argument in the path.
$B$ is multifold connected to $A$ because two paths connecting $B$ to $A$ exist.
\end{example}

Next, we %explain 
give some qualitative and quantitative guarantees for AAEs that will be useful to prove
%interesting 
properties of AAEs later.

% We start with direct connectivity as it represents straight influence from one argument to another. 
We start by showing that AAEs correctly capture the
qualitative effect (positive or negative)
of direct connectivity, in that  
arguments attacking (supporting) the topic argument always have negative (positive, respectively) or zero attribution scores.

\begin{proposition}[Direct Qualitative Attribution Influence]
\label{prop_point_qua}
If $B, A \in \mathcal{A}$ are directly connected, then
    \begin{enumerate}
        \item If $(B, A) \in \mathcal{R^{-}}$, then $\left. \nabla \right|_{B \mapsto A} \leq 0$;
        \item If $(B, A) \in \mathcal{R^{+}}$, then $\left. \nabla \right|_{B \mapsto A} \geq 0$.
    \end{enumerate}
\end{proposition}
%\begin{proof}
%See Appendix.
%\end{proof}

% \todo{where is the appendix? :)}
% \todo{not yet finished lmao}

% \begin{proof}
% $$
% \begin{aligned}
%     \left. \nabla \right|_{B \mapsto A} 
%     &= \lim_{\varepsilon \to 0} \frac{f_{att}(v_{B}+\varepsilon)-f_{att}(v_{B})}{(v_{B}+\varepsilon)-v_{B}}\\
%     &= \frac{v_{A} \cdot \left[ 1-(v_{B}+\varepsilon) \right] - v_{A} \cdot \left[ 1-v_{B} \right] }{(v_{B}+\varepsilon)-v_{B}}
% \end{aligned}
% $$
% where numerator $v_{A} \cdot \left[ 1-(v_{B}+\varepsilon) \right] - v_{A} \cdot \left[ 1-v_{B} \right] < 0$ and denominator$(v_{B}+\varepsilon)-v_{B} > 0$, therefore proportion $\left. \nabla \right|_{B \mapsto A}<0$
% \end{proof}

% From proposition \ref{prop_point_qua}, we know that attacker argument has a negative influence on its direct adjacent argument; while supporter argument has a positive influence on its direct adjacent argument.

% \todo{We illustrate ...}

%By Definition~\ref{def_attribution},
The next proposition
gives an exact quantification of the influence of arguments $B$ %, in the case, 
directly attacking or supporting the topic argument $A$, in terms of
%concerns analytical quantitative influence. 
%The attribution from $B$ to $A$ depends upon 
three parameters: the base score of $A$, the aggregation strength of $B$ and the strength of other arguments $Z$ that are in the same relation with $A$ as $B$% as $(B,A)$
.
\begin{proposition}[Direct Quantitative Attribution Influence]
\label{prop_point_quan}
If $B, A \in \mathcal{A}$
are directly connected and $(B,A) \in \mathcal{R^{*}}$%\todo{what is this?}\todo[inline]{Xiang: R* is either R- or R+. If (B,A) is R-, then (Z,A) is R-, too. We use R* to denote (B,A) and (Z,A) share the same relation} 
, for $\mathcal{R^{*}}=\mathcal{R^{-}}$ or $\mathcal{R^{*}}=\mathcal{R^{+}}$, then
$$
\left. \nabla \right|_{B \mapsto A}=\xi_{B}  (1-\left| v_{Ba}-v_{Bs} \right|)  \prod_{\left \{ Z \in \mathcal{A}\setminus{B} \mid (Z,A) \in \mathcal{R^{*}} \right \} }\left[ 1-\sigma(Z) \right],
$$
where $$
\xi_{B}=    
    \begin{cases}
        -\tau(A) & if \ \ \mathcal{R^{*}}=\mathcal{R^{-}} \wedge v_{Aa} \geq v_{As};\\
        (\tau(A)-1) & if \ \ \mathcal{R^{*}}=\mathcal{R^{-}} \wedge v_{Aa} < v_{As};\\
        \tau(A) & if \ \ \mathcal{R^{*}}=\mathcal{R^{+}} \wedge v_{Aa} > v_{As};\\
        (1-\tau(A)) & if \ \ \mathcal{R^{*}}=\mathcal{R^{+}} \wedge v_{Aa} \leq v_{As}.\\
    \end{cases}
$$
\end{proposition}

Inspired by the chain rule of gradients, we further study some qualitative and quantitative guarantees of indirect connectivity in the next two propositions. 
First, we find that one argument indirectly connected to a topic argument always has positive (negative) or neutral influence if the number of attacks in the path between the argument and the topic argument is even (odd, respectively).
% The next two propositions make analogous statements for indirect connectivity. 

% \todo{simplify Prop. 3 and 4 similar to Prop. 1 and 2 (delete "QBAF Q..." and other redundant notation)}
\begin{proposition}[Indirect Qualitative Attribution Influence]
\label{prop_path_qua}
% Given an acyclic QBAF $\mathcal{Q}=\left\langle\mathcal{A}, \mathcal{R}^{-}, \mathcal{R}^{+}, \tau \right\rangle$, $X_{1},\ldots, X_{n} \in \mathcal{A}$.
% and $\mathcal{R} = \mathcal{R}^{-} \cup \mathcal{R}^{+}$. 
Let $X_{1},\ldots, X_{n} \in \mathcal{A}$ ($n \geq 3$) and $\mathcal{R} = \mathcal{R}^{-} \cup \mathcal{R}^{+}$.
% and $\Phi_{X_{1} \mapsto X_{n}}=\{\phi\}$.
Suppose $S=\{(X_{1}, X_{2}), (X_{2}, X_{3}),\ldots, (X_{n-1}, X_{n})\} \subseteq \mathcal{R}$. Let $|S \cap \mathcal{R}^{-}|=\Theta$.
If $X_{1},X_{n}\in \mathcal{A}$ are indirectly connected through path $\phi=X_{1},\ldots, X_{n}$, then
    \begin{enumerate}
        \item If $\Theta$ is odd, then $\left. \nabla%_{\phi} 
        \right|
        _{X_{1} \mapsto X_{n}} \leq 0$;
        \item If $\Theta$ is even, then $\left. \nabla%_{\phi} 
        \right|_{X_{1} \mapsto X_{n}} \geq 0$.
    \end{enumerate}
\end{proposition}
% \begin{proof}
% See Appendix.
% \end{proof}
%The next proposition is analogous to the chain rule which states that 
%According to Definition~\ref{def_attribution}, 
Then we show that the AAE %attribution score 
from one argument $X_1$ to an indirectly connected topic argument $X_n$ can be precisely characterized in terms of 
%equals to the product of these attribution scores divided by some parameters w.r.t. their aggregation strength value $v$.
the attribution scores from the arguments in the path from $X_1$ to $X_n$ and the strengths of these arguments.  
\begin{proposition}[Indirect Quantitative Attribution Influence]
\label{prop_path_quan}
Let $X_{1},\ldots, X_{n} \in \mathcal{A}$. If $X_{1},X_{n}\in \mathcal{A}$ are indirectly connected through path $\phi=X_{1},\ldots, X_{n}$, then
% Given an acyclic QBAF $\mathcal{Q}=\left\langle\mathcal{A}, \mathcal{R}^{-}, \mathcal{R}^{+}, \tau \right\rangle$, $X_{1},\ldots, X_{n} \in \mathcal{A}$. %and $\mathcal{R} = \mathcal{R}^{-} \cup \mathcal{R}^{+}$. 
% Let $X_{1}$ and $X_{n}$ be indirectly connected such that $\phi=X_{1},\ldots, X_{n}$ and $\Phi_{X_{1} \mapsto X_{n}}=\{\phi\}$. Then
$$
\left. \nabla \right|_{X_{1} \mapsto X_{n}}=(1-\left| v_{X_{1}{a}}-v_{X_{1}{s}} \right|) \cdot \prod_{i=1}^{n-1}\frac{\left. \nabla \right|_{X_{i} \mapsto X_{i+1}}}{(1-\left| v_{X_{i}{a}}-v_{X_{i}{s}} \right|)}.
$$
\end{proposition}
We %show an example of 
illustrate next the application of the propositions %mentioned 
in this section %to compute the attribution in Example \ref{example_QBAF_abc}
% (see \url{https://arxiv.org/abs/2307.13582} for more examples).
(see the SM for more examples).  %%% ARXIV VERSION
\begin{example}
\label{example_QBAF_abc}
% Consider an acyclic QBAF $\mathcal{Q}=\left\langle\mathcal{A}, \mathcal{R}^{-}, \mathcal{R}^{+}, \tau \right\rangle$ equipped with DF-QuAD gradual semantics $\sigma$ in Figure \ref{fig_bcda_two_path}.
% Suppose the base score $\tau$ of every argument is 0.5 in $\mathcal{Q}$.\\
The settings are the same as in Example \ref{example_DF_QuAD}.
According to Proposition \ref{prop_point_qua}, we have $\left. \nabla \right|_{D \mapsto A} \leq 0$ because $(D,A)\in\mathcal{R^{-}}$.
According to Proposition \ref{prop_point_quan}, we have
$$
\begin{aligned}
    \left. \nabla \right|_{D \mapsto A}
    &= -\tau(A) \cdot (1-\left| v_{Da}-v_{Ds} \right|)\\
    &\ \cdot \prod_{\left \{ Z \in \mathcal{A}\setminus{D} \mid (Z,A) \in \mathcal{R^{-}} \right \} }(1-\sigma(Z))=-0.375.\\
\end{aligned}
$$
Similarly, $\left. \nabla \right|_{C \mapsto D}=-0.25$.
According to Proposition \ref{prop_path_qua}, we have $\left. \nabla \right|_{C \mapsto A} \geq 0$ because the number of attacks %relations 
is two (even).
According to Proposition \ref{prop_path_quan}, we have
$$
\begin{aligned}
    \left. \nabla \right|_{C \mapsto A}
    &= \left. \nabla \right|_{C \mapsto D} \times \frac{\left. \nabla \right|_{D \mapsto A}}{1-\left| v_{Da}-v_{Ds} \right|}=0.125.
\end{aligned}
$$
% Similarly, $\left. \nabla_{\phi} \right|_{B \mapsto C \mapsto D \mapsto A}=-0.125$.

% According to Proposition \ref{prop_graph_wise_influence}, 
% $$
% \begin{aligned}
% \left. \nabla_{\Phi} \right|_{B \mapsto A}&=\sum_{\phi \in \Phi_{B \mapsto A}}\left. \nabla_{\phi} \right|_{B \mapsto A}=-0.25\\
% \end{aligned}
% $$
\end{example}
\section{Properties}
\label{sec_properties}
%\todo[inline]{NP: I would replace iff (if and only if) with if in the theorems. If you really considered all possible cases, iff may actually be accurate, but since you probably do not explain the only-part in the proofs, I would just write if to avoid technical problems}
We now study some desirable properties~\footnote{With `desirable' we mean properties that a user may demand from an argumentative explanation method.} of our AAEs %that explain 
as explanations for
outcomes under the DF-QuAD gradual semantics. 
Although the properties that AAEs can and should satisfy depend, of course, on the underlying semantics, our properties make guarantees about the explanations, not about the underlying semantics. %Here, we particularly focus on the \emph{faithfulness} and low \emph{computational complexity} of our explanations.

We start with the \emph{explainability} property
from \cite{amgoud2022axiomatic},  which guarantees that an explanation always exist.
This property, for AAEs, amounts to well-definedness, and can be formulated as follows.
\begin{proposition}[Explainability]
\label{proposition_explainability}
$\forall A,B \in \mathcal{A}$,
$
\left. \nabla \right|_{B \mapsto A} \in \mathbb{R}
$ is well-defined.
\end{proposition}

\emph{Missingness} is an important property for attribution methods,  guaranteeing the \emph{faithfulness} of the explanation for non-relevant features \cite{lundberg2017unified}. Here, it states that if one argument is not connected to the topic argument, then the AAE %attribution score %would be 
is zero.

\begin{proposition}[Missingness]
\label{proposition_missingness}
$\forall A, B \in \mathcal{A}$, if $B$ is %not connected to
disconnected 
from $A$, then
$$
\left. \nabla \right|_{B \mapsto A} = 0.
$$
\end{proposition}

The next four properties guarantee the \emph{faithfulness} of AAEs to the underlying QBAF. The former two consider the faithfulness of one particular argument while the latter two consider the faithfulness between arguments.

We propose \emph{completeness}, inspired by \cite{shrikumar2017learning} and by \emph{quantitative faithfulness} in \cite{potyka2022towards}. In our %AAE
setting, it states that 
the change of the strength of the topic argument should be proportional to 
%the attribution score of the argument
its AAE.
\begin{property}[Completeness]
\label{property_completeness}
Let $A, B \in \mathcal{A}$ and let $\sigma'_B(A)$ denote the strength of $A$ when setting $\tau(B)$ to $0$. Then 
$$
   -\tau(B) \cdot \left. \nabla \right|_{B \mapsto A} = \sigma_{B}'(A)-\sigma(A).
$$
\end{property}

\begin{proposition}
\label{proposition_completeness_satisfy}
Completeness is satisfied if $B$ is directly or indirectly connected to $A$.
\end{proposition}
\begin{proposition}
\label{proposition_completeness_violate}
Completeness can be violated if $B$ is multifold connected to $A$.
\end{proposition}

\emph{Counterfactuality} is inspired by \cite{covert2021explaining}, which considers the situation of removing arguments.
This property states that if one argument $B$ has a positive (negative) influence on $A$, then removing $B$ will decrease (increase, respectively) the strength of $A$. 
\begin{property}[Counterfactuality]
\label{property_counterfactuality}
Let $A, B \in \mathcal{A}$ and let $\sigma'_B(A)$ denote the strength of $A$ when setting $\tau(B)$ to $0$. Then
\begin{enumerate}
    \item If $\left. \nabla \right|_{B \mapsto A} \leq 0$, then $\sigma_{B}'(A) \geq \sigma(A)$;
    \item If $\left. \nabla \right|_{B \mapsto A} \geq 0$, then $\sigma_{B}'(A) \leq \sigma(A)$.
\end{enumerate}
\end{property}

\begin{proposition}
\label{proposition_counterfactuality_satisfy}
Counterfactuality is satisfied if $B$ is directly or indirectly connected to $A$.
\end{proposition}
\begin{proposition}
\label{proposition_counterfactuality_violate}
Counterfactuality can be violated if $B$ is multifold connected to $A$.
\end{proposition}

We propose \emph{agreement} as a property for comparing %attribution scores %between 
AAEs
across any two arguments.
It states that two arguments have the same influence on the strength of a topic argument whenever they have the same base scores and the same %attribution scores.
AAEs to the topic argument.
% their base and attribution scores are equal or their proportions are, i.e., $\frac{|\tau(B)|}{|\tau(C)|} = \frac{|\nabla|_{B \mapsto A}|}{|\nabla|_{C \mapsto A}|}$.

% Explanations that fail to satisfy this property are not faithful explanations.
\begin{property}[Agreement]
\label{property_agreement}
Let $A, B, C \in \mathcal{A}$ and $\sigma'_B(A),\sigma'_C(A)$ denote the strength of $A$ when setting $\tau(B),\tau(C)$ to $0$ respectively.\\
If $$\left| \tau(B) \cdot \left. \nabla \right|_{B \mapsto A} \right| = \left| \tau(C) \cdot \left. \nabla \right|_{C \mapsto A} \right|$$
then $$\left| \sigma_{B}'(A)-\sigma(A) \right| = \left| \sigma_{C}'(A)-\sigma(A) \right|.$$
\end{property}

\begin{proposition}
\label{proposition_agreement_satisfy}
Agreement is satisfied if $B$ and $C$ are directly or indirectly connected to $A$.
\end{proposition}
\begin{proposition}
\label{proposition_agreement_violate}
Agreement can be violated if $B$ or $C$ is multifold connected to $A$.
\end{proposition}

\emph{Monotonicity} states that features with larger attribution scores have  a larger influence on the output \cite{chen2022makes}. In our setting, monotonicity means that the larger the AAE %contribution of the 
from an argument, the more influence it should have on the strength of the topic argument.
\begin{property}[Monotonicity]
\label{property_monotonicity}
Let $A, B, C \in \mathcal{A}$, $\sigma'_B(A),\sigma'_C(A)$ denote the strength of $A$ when setting $\tau(B),\tau(C)$ to $0$ respectively.\\
If $$\left| \tau(B) \cdot \left. \nabla \right|_{B \mapsto A} \right| < \left| \tau(C) \cdot \left. \nabla \right|_{C \mapsto A} \right|$$ 
then $$\left| \sigma_{B}'(A)-\sigma(A) \right| < \left| \sigma_{C}'(A)-\sigma(A) \right|.$$
\end{property}

\begin{proposition}
\label{proposition_monotonicity_satisfy}
Monotonicity is satisfied if $B$ and $C$ are directly or indirectly connected to $A$.
\end{proposition}
\begin{proposition}
\label{proposition_monotonicity_violate}
Monotonicity can be violated if $B$ or $C$ is multifold connected to $A$.
\end{proposition}
% \todo[inline]{I have added the intuition why we study Invariability}
Although gradient-based attributions such as AAEs are \emph{local} rather than \emph{global} explanations in general, we %indeed
find some interesting qualitative and quantitative global guarantees for AAEs under direct and indirect connectivity% settings
.
\emph{Qualitative invariability} states that one argument will always have a positive (or negative) influence on the topic arguments, regardless of the change of its base score, while \emph{quantitative invariability} shows that the attribution score of one argument will keep invariant with the change of its base score.
\begin{property}[Qualitative Invariability]
\label{property_qua_invar}
$\forall A, B \in \mathcal{A}$, let $\nabla_{\delta}$ denote the AAE from $B$ to $A$ when setting $\tau(B)$ to some $\delta \in [0,1]$. Then
\begin{enumerate}
    \item If $\left. \nabla \right|_{B \mapsto A} \leq 0$, then $\forall \delta \in [0,1]$, $\nabla_{\delta} \leq 0$;
    \item If $\left. \nabla \right|_{B \mapsto A} \geq 0$, then $\forall \delta \in [0,1]$, $\nabla_{\delta} \geq 0$.
\end{enumerate}
\end{property}

\begin{proposition}
\label{proposition_qua_invar_satisfy}
Qualitative invariability is satisfied if $B$ is directly or indirectly connected to $A$.
\end{proposition}
\begin{proposition}
\label{proposition_qua_invar_violate}
Qualitative invariability can be violated if $B$ is multifold connected to $A$.
\end{proposition}

\begin{property}[Quantitative Invariability]
\label{property_quan_invar}
$\forall A, B \in \mathcal{A}$, %$C \in \mathbb{R}$, 
let $\nabla_{\delta}$ denote the AAE from $B$ to $A$ when setting $\tau(B)$ to some $\delta \in [0,1]$.
For all $\delta$, there always exists a constant $C \in [-1,1]$ such that
$$
\nabla_{\delta} \equiv C.
$$
\end{property}

\begin{proposition}
\label{proposition_quan_invar_satisfy}
Quantitative invariability is satisfied if $B$ is directly or indirectly connected to $A$.
\end{proposition}
\begin{proposition}
\label{proposition_quan_invar_violate}
Quantitative invariability can be violated if $B$ is multifold connected to $A$.
\end{proposition}

The final property is \emph{tractability}, measuring the computational complexity of AAEs. This property shows that our explanation method is efficient in the sense that the AAEs %of the arguments
can be generated in linear time in the number of arguments.
\begin{proposition}[Tractability]
\label{proposition_tractability}
If $\left| \mathcal{A} \right|=n$, then AAEs can be generated in linear time $\mathcal{O}(n)$.
\end{proposition}
% \todo[inline]{I have added a further discussion about multifold connectivity, and added a footnote}
% Here, we further analyse why 
Let us note that many of the aforementioned properties
do not hold for the multifold connectivity case.
% Since our properties make guarantees about explanations rather than semantics, therefore, it is difficult to guarantee the properties of explanations if DF-QuAD does not satisfy some properties of semantics.
We would like to point out that this is a feature, not a bug, for an explanation
method that is supposed to be faithful to the underlying argumentation semantics.
This is because many properties satisfied by DF-QuAD, like \emph{balance} and \emph{monotonicity}\footnote{Here monotonicity refers to guarantees about DF-QuAD, and is different from Property~\ref{property_monotonicity}, which gives guarantees about the explanations.}~\cite{baroni2018many}, make only guarantees about the effect 
of direct attackers or supporters on an argument. The effect in
the multifold connectivity case depends on various other factors (including base
scores of all directly and indirectly connected arguments) and there is an
infinite (finite for a fixed number of involved arguments) number of special cases that may occur. 

% , which does not guarantee multifold cases.
% Although multifold connectivity does not hold for some properties, it does not affect the applicability of AAEs because we focus on the ranking of relative attribution scores of arguments.

\section{Case Studies}
\label{sec_cases}
We carry out two case studies to show the applicability and usefulness of AAEs in %fake news detection and movie recommender systems
practice% as demonstrations
.
% (see the SM for an analysis of the fraud detection problem discussed in the Introduction).

\begin{figure}[ht]
	\centering
		\includegraphics[width=0.58\columnwidth]{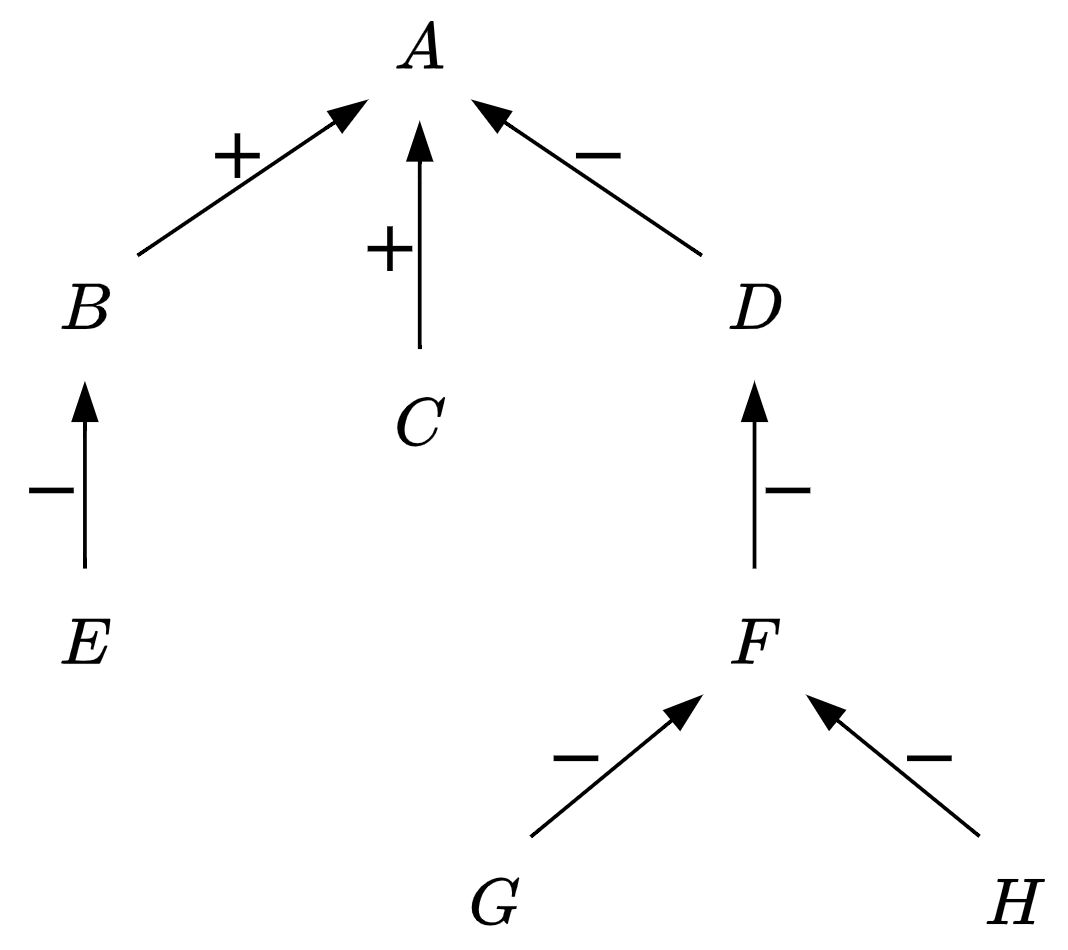}
	\caption{Structure of a QBAF for fake news detection (taken from \cite{kotonya2019gradual}).}
	\label{fig_Neema}
\end{figure}

\textbf{Case Study 1: Fake News Detection.}\\
Fake news detection plays an important role in avoiding the spread of rumors on social media.
In \cite{kotonya2019gradual}, QBAFs are used to detect whether a source tweet is a rumor by aggregating the weight of replies to the source tweet.
Concretely, Figure \ref{fig_Neema} from \cite{kotonya2019gradual} shows an intuitive example of using QBAFs to detect a rumor tweet. The QBAF in the figure is built up by extracting the dialectical relation between the source tweet and the replies with the help of deep learning techniques. The %conversation 
thread of tweets w.r.t Figure \ref{fig_Neema} is as follows.\\
\emph{\boldmath{$A$} [\textbf{u1/source tweet}] Up to 20 held hostage in Sydney Lindt Cafe siege.}\\
\emph{\boldmath{$B$} [\textbf{u2/reply1}] @u1 pretty sure it was}\\
\emph{\boldmath{$C$} [\textbf{u3/reply2}] @u1 yeah terrible}\\
\emph{\boldmath{$D$} [\textbf{u4/reply3}] @u1 all reports say 13}\\
\emph{\boldmath{$E$} [\textbf{u5/reply4}] @u2 nonsense}\\
\emph{\boldmath{$F$} [\textbf{u6/reply5}] @u4 this number is ridiculous}\\
\emph{\boldmath{$G$} [\textbf{u7/reply6}] @u6 not convincing at all}\\
\emph{\boldmath{$H$} [\textbf{u8/reply7}] @u6 you are an insensitive idiot!}\\

In Figure \ref{fig_Neema}, the task is to detect whether $A$ is a rumor, hence $A$ is the topic argument.
$B$, $C$ and $D$ are the direct replies for $A$. $E$ is the reply for $B$; $F$ is the reply for $D$; $G$ and $H$ are replies for $F$. Therefore, $E$, $F$, $G$ and $H$ are indirect replies for $A$.
Since the authors of \cite{kotonya2019gradual}  mainly focus on the overall performance for fake news detection, they do not provide %a particular conversation thread 
explanations 
with respect to this QBAF. Here, we show possible %corresponding 
conversation threads as qualitative explanations which are consistent with the intuition of \cite{kotonya2019gradual}, before showing how AAEs can additional provide quantitative explanations.

The base scores for all the arguments are initially set to 0.5. According to the DF-QuAD gradual semantics, the strength for $A$ is $\sigma(A)=0.59375$, which is seen as ``True News" because $\sigma(A)>0.5$.
Instead of just providing this prediction, %QBAFs are also able to generate 
qualitative %dialogical
conversational explanations can be obtained from %their own 
the QBAF structure.
For instance, an explanation for the output decision might be:

\begin{quote}
\emph{\textbf{User}: Why the source tweet $A$ is not a rumor?}\\
\emph{\textbf{QBAF}: Because replies $B$ and $C$ support $A$, despite $D$ is against $A$.}\\
\emph{\textbf{User}: Why $A$ is still true in spite of the attack of $D$?}\\
\emph{\textbf{QBAF}: Although $D$ attacks $A$, $D$ is also attacked by $F$, which decreases $D$'s strength when attacking $A$. Although $G$ and $H$ attack $F$, $F$ still supports $A$, just with a weaker strength.}    
\end{quote}

% In this interactive explanations, argumentative explanations are extracted directly from the structure of this QBAF. 
% \todo[inline]{I have stressed the advantage of our AAE compared with dialogical explanations}
However, %based on dialogical
using these conversational explanations alone, it is %still
unclear how much %quantitative attribution of 
each argument %leads 
contributes 
to the final strength of the topic argument. Next, we apply AAEs to %quantitatively 
explain the outcome %reasoning 
for $A$%, also demonstrating results visually
.

%According to Definition~\ref{def_attribution}, w

We compute the AAEs in Table~\ref{strength_gradient} (last column), presented in descending order to obtain a ranking. These attributions %are also able to 
can also be computed by applying Propositions \ref{prop_point_quan} and \ref{prop_path_quan}, because all arguments (except $A$) are either directly or indirectly connected to $A$. Based on the ranking, we can obtain qualitative and quantitative analyses that %dialogical 
qualitative explanations alone are unable to provide, as follows.
\begin{table}
\caption{AAEs in descending order (last column) for the QBAF in Figure \ref{fig_Neema}.}
% \vskip -0.25in
\label{strength_gradient}
\begin{center}
\begin{small}
\begin{sc}
\begin{tabular}{ccccc}
\toprule
Argument & $\tau$& $\sigma_X'(A)$& $\sigma_X'(A)-\sigma(A)$& $\nabla$\\
\midrule
Reply2: C  &0.5 &0.40625 &-0.18750 &0.3750\\
Reply1: B  &0.5 &0.53125 &-0.06250 &0.1250\\
Reply5: F  &0.5 &0.56250 &-0.03125 &0.0625\\
Reply6: G  &0.5 &0.62500 &0.03125 &-0.0625\\
Reply7: H  &0.5 &0.62500 &0.03125 &-0.0625\\
Reply4: E  &0.5 &0.65625 &0.06250 &-0.1250\\
Reply3: D  &0.5 &0.81250 &0.21875 &-0.4375\\
\bottomrule
\end{tabular}
% \captionsetup{skip=5pt}
\end{sc}
\end{small}
\end{center}
% \vskip -0.35in
\end{table}

\paragraph{Qualitative Analysis}
$B$, $C$ and $F$ have a positive influence on $A$. Indeed $B$ and $C$ directly support $A$, and a path with two attackers link $F$ to $A$, so $F$ has a positive influence on $A$ as well. 
$D$, $E$, $G$ and $H$ have a negative influence on $A$. $D$ directly attacks $A$, while there is an  odd number of attackers from $E$, $G$ and $H$ to %they have the  relations in the path from themselves to 
$A$. This qualitative analysis complements well the qualitative, conversational explanation we saw earlier.
% \begin{table}
% \begin{center}
% {\caption{Argument attributions in descending order for Figure \ref{fig_Neema}.}\label{strength_gradient}}
% \begin{tabular}{ccccc}
% \hline
% \rule{0pt}{12pt}
% Argument & $\tau$& $\sigma_X'(A)$& $\sigma_X'(A)-\sigma(A)$& $\nabla$\\
% \hline
% \\[-4pt]
% Relpy2: C  &0.5 &0.40625 &-0.18750 &0.3750\\
% Relpy1: B  &0.5 &0.53125 &-0.06250 &0.1250\\
% Relpy5: F  &0.5 &0.56250 &-0.03125 &0.0625\\
% Relpy6: G  &0.5 &0.62500 &0.03125 &-0.0625\\
% Relpy7: H  &0.5 &0.62500 &0.03125 &-0.0625\\
% Relpy4: E  &0.5 &0.65625 &0.06250 &-0.1250\\
% Relpy3: D  &0.5 &0.81250 &0.21875 &-0.4375\\
% \hline
% \\[-6pt]

% % \multicolumn{8}{l}{$\Diamond$ execution time in ticks\ \
% % $\Box$ speed-up values\ \
% % $\bigtriangleup$ efficiency values}
% \end{tabular}
% \end{center}
% \end{table}

\paragraph{Quantitative Analysis}
Among  arguments with a positive attribution influence, $C$ has the largest influence on $A$, while the influence of $B$ is less than $C$ because argument $E$ attacks $B$, thus weakening $B$'s influence.  $F$ has the smallest positive influence on $A$ because the influence is indirect, and, at the same time, two attackers of $F$ weaken its positive influence.
Of all the arguments with a negative attribution influence, $D$ has the highest influence on $A$. This is because $D$  not only directly attacks $A$, but also has indirect supporters from $G$ and $H$ as they attack $F$, the attacker of $D$.
$E$ has an indirect negative influence on $A$ by attacking its supporter $B$. Then, $G$ and $H$ have a smaller negative influence than $B$ since they are farther away than $B$.
% \todo[inline]{I have stressed the advantage of AAEs}

Additionally, note that when the base scores of arguments change, the attribution ranking changes accordingly, giving rise to different explanations. 
However, qualitative explanations always remain identical regardless of the numerical information conveyed by AAEs, because they only focus on the structure of the QBAFs.

\begin{figure}[t]
	\centering
		\includegraphics[width=1.0\columnwidth]{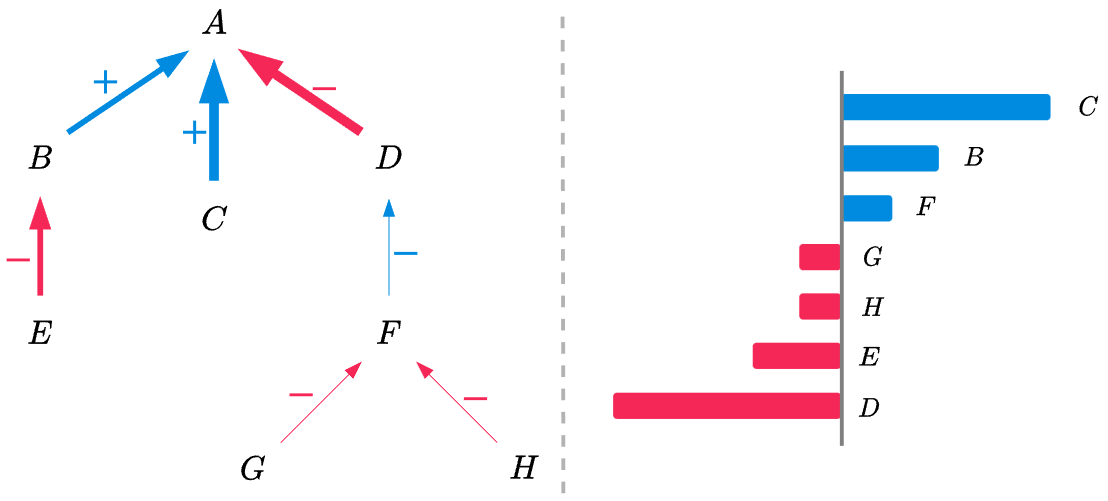}
	\caption{Visualization of AAEs for fake news detection.}
	\label{fig_rank_fakenews}
\end{figure}

\vspace*{-0.2cm}
\paragraph{Visualization}
In order to be provide intuitive %and understandable 
explanations, we can visualize the AAEs for instance as shown in Figure \ref{fig_rank_fakenews}.
The left-hand side shows the AAEs to the topic argument. Here, blue arrows show  positive attribution while red arrows show negative attribution, and the thickness of arrows shows the magnitude of the attribution.
The right-hand side shows ranking and magnitude of the AAEs.

\paragraph{Property Analysis}
% \todo[inline]{I have added reasons why we do property analysis}
We %verify 
discuss the satisfaction of the proposed properties to show that AAEs are a good explanation method% evaluated by several desirable properties
.
In Table \ref{strength_gradient}, we can see that each argument is assigned a real number as attribution to the topic argument, which satisfies \emph{explainability}.
%For any argument like $B$, 
If we change the base score of $B$ in the range of $[0,1]$, the attribution score will not change ($\left. \nabla \right|_{B \mapsto A}=0.125>0$), which satisfies \emph{invariability} qualitatively and quantitatively.
Next, we demonstrate the \emph{faithfulness} of AAEs by analyzing \emph{missingness}, \emph{counterfactuality}, \emph{completeness}, \emph{agreement} and \emph{monotonicity}.
For any two arguments without any connections like $C$ and $E$, the attribution score is $0$ ($\left. \nabla \right|_{C \mapsto E}=0$), which is intuitive and satisfies \emph{missingness}.
For any argument whose attribution score is positive like $C$, then setting the base score %of this argument 
to $0$ will decrease the strength of $A$ from $0.59375$ to $0.40625$, and $\sigma_C'(A)-\sigma(A)=-\tau(C) \cdot \left. \nabla \right|_{C \mapsto A}=-0.1875$,
which satisfies both \emph{counterfactuality} and \emph{completeness}. 
Any two arguments with the same contributions, like $G$ and $H$,  have the same influence on $A$. Given that $\left| \tau(G) \cdot \left. \nabla \right|_{G \mapsto A} \right| = \left| \tau(H) \cdot \left. \nabla \right|_{H \mapsto A} \right|=0.03125$, we have $\left| \sigma_{G}'(A)-\sigma(A) \right| = \left| \sigma_{H}'(A)-\sigma(A) \right|$, which satisfies \emph{agreement}. 
For those with different contributions, like $B$ and $C$, due to $\left| \tau(B) \cdot \left. \nabla \right|_{B \mapsto A} \right|=0.0625 < \left| \tau(C) \cdot \left. \nabla \right|_{C \mapsto A} \right|=0.1875$, we get $\left| \sigma_{B}'(A)-\sigma(A) \right|\!=\!0.0625 \!< \! \left| \sigma_{C}'(A)-\sigma(A) \right|\!=\!0.1875$, guaranteeing  \emph{monotonicity}.
From the computational angle, all AAEs can be computed in linear time in the number of arguments% in QBAF
.
\\
% \textbf{Monotonicity}
% Let us consider monotonicity in this argumentation. For all the arguments except for $A$, they are either point-wise or path-wise linked to $A$, therefore, all these arguments are globally monotonic to $A$. That is, if the base score of one arguments changed (while keeping the others still), then the influence will not change.

\begin{figure}[t]
	\centering
	\includegraphics[width=0.55\columnwidth]{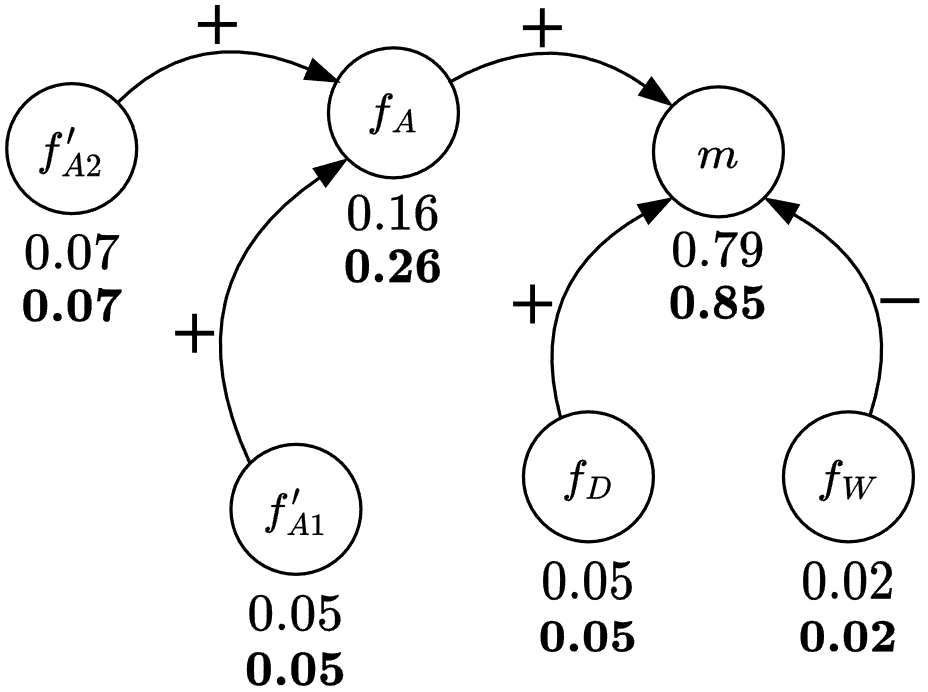}
	\caption{%Structure of a 
 QBAF for movie recommendations (taken from \cite{cocarascu2019extracting}). Values in regular font are base scores and  values in bold are DF-QuAD strengths.}
	\label{fig_proof_prop6_exp}
\end{figure}

\textbf{Case Study 2: Movie Recommender Systems.}\\
Online review aggregation has become an increasingly effective quality control method in the era of information explosion.
\cite{cocarascu2019extracting} propose to build up movie recommender systems based on QBAFs by aggregating online movie reviews.
Basically, they build  Argumentative Dialogical Agents (ADAs) to aggregate movie reviews in natural language and then generate  ratings of movies from the aggregations, recommending highly rated movies to users.
ADAs can also engage in interactive explanations (conversations) with users explaining why recommended movies are highly rated, leveraging on the underlying QBAFs. Let us take movie \emph{The Post} as an example in their paper. 
%First, an ADA is extracted from  snippets of reviews (using NLP techniques), where movie %rating is aggregated from four features: \emph{acting}, \emph{directing}, \emph{themes} and \emph{writing}. 
Figure \ref{fig_proof_prop6_exp} shows the QBAF underpinning an ADA for this example,  
% Then, applying the DF-QuAD algorithm to compute the rating of the movie. The normal font values are the base score and the bold font values are the strength of arguments. In Figure \ref{fig_proof_prop6_exp}, 
where the topic argument $m$ stands for the movie in question and %four
three
features support or attack $m$ (here
$f_A$ stands for \emph{acting},
$f_D$ stands for \emph{directing}, and $f_W$ stands for \emph{writing}). %Based on NLP techniques, no relations are mined from the $f_T$ hence $f_T$ is not shown in Figure \ref{fig_proof_prop6_exp}.
% has neutral influence on $m$, which is denoted in dashed arrow. 
%
%Each feature may have sub-features if it is mined by NLP techniques. 
Some of these features have sub-features ($f_A$ has sub-features $f'_{A1}$ and $f'_{A2}$).
The base scores in the QBAF result from the review aggregation method employed by ADA.
%After building up the ADA, we can see the final rating of the movie is 0.85, 
ADA then applies DF-QuAD to this QBAF, to obtain the argument strengths indicated in bold in Figure~\ref{fig_proof_prop6_exp}. 
A possible %dialogue
conversational explanation for this example is shown as follows in \cite{cocarascu2019extracting}.

\begin{quote}
\emph{\textbf{User}: Why was The Post highly rated?}\\
\emph{\textbf{ADA}: This is because the acting was really great, although the writing was a little poor.}\\
\emph{\textbf{User}: Why was the acting great?}\\
\emph{\textbf{ADA}: Because actor Meryl Streep was great.}\\
\emph{\textbf{User}: What did critics say about Meryl Streep being great?}\\
\emph{\textbf{ADA}: “...Streep’s hesitations, %rue, and ultimate valor 
... are soul-deep...”}  
\end{quote}

\begin{figure}[t]
	\centering
		\includegraphics[width=1.0\columnwidth]{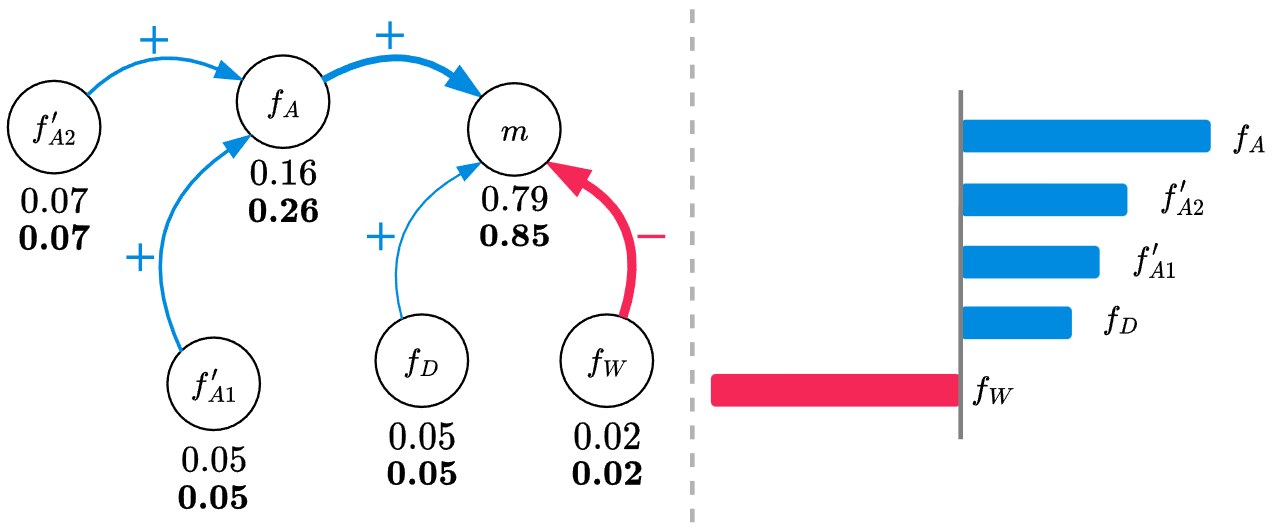}
	\caption{%AAEs for 
 Visualization of AAEs for movie recommendations in ADA.}
	\label{fig_rank_movie}
\end{figure}

% Next, we will apply the argument attribution explanation method to further explain the in this ADA. 
As before, we can apply AAEs to further explain $m$ in this %ADA.
example. We give the AAEs %ranking of the argument attributions is shown 
in Table~\ref{tab_recommender_system} and visualizations in Figure~\ref{fig_rank_movie}.

\begin{table}
\caption{%Argument attributions in descending order for 
AAEs in descending order (last column) for the QBAF in Figure \ref{fig_proof_prop6_exp}.}
% \vskip -0.3in
\label{tab_recommender_system}
\begin{center}
\begin{small}
\begin{sc}
\begin{tabular}{ccccc}
\toprule
Argument & $\tau$& $\sigma_X'(m)$& $\sigma_X'(m)-\sigma(m)$& $\nabla$\\
\midrule
Acting     &0.16 &0.81954 &-0.02820 &0.17625\\
Actor2     &0.07 &0.83660 &-0.01114 &0.15920\\
Actor1     &0.05 &0.83995 &-0.00779 &0.15585\\
Directing  &0.05 &0.83995 &-0.00779 &0.15584\\
Writing    &0.02 &0.85194 &0.0042  &-0.21\\
\bottomrule
\end{tabular}
\end{sc}
\end{small}
\end{center}
% \vskip -0.3in
\end{table}

% The analysis of connectivity, qualitative influence and quantitative influence are omitted since they are much analogous to those in case study 1. Here, we focus more on the \emph{faithfulness} of our method. In the recent literature, \cite{nguyen2020quantitative} propose to use spearsman correlation ........

% \paragraph{Connectivity}
% $m$ is the topic argument, which represents the movie \emph{The Post}. 
% According to Definition \ref{def_connectivity},
% $f_{A}$, $f_{D}$, $f_{T}$ and $f_{W}$ are four \emph{direct} feature-based characterisation of $m$, the connectivity between them and $m$ are direct connectivity; $f'_{A1}$ and $f'_{A2}$ are two sub-features for $f_{A}$, indirect characterisations of $m$, which are indirectly connected to $m$.

\paragraph{Qualitative Analysis}
According to Propositions \ref{prop_point_qua} and \ref{prop_path_qua},
$f_{A}$ and $f_{D}$ have a positive influence on $m$ because they are supporters% for $m$
; $f_{W}$ has a negative influence on $m$ because it is an attacker of $m$.
% ; $f_{T}$ has neutral influence on $m$, which means this QBAF does not mine some relation between themes and the movie.
$f'_{A1}$ and $f'_{A2}$ both have positive influence on $m$, because they are supporters of a supporter of $m$, and 0 (even) attackers  from them to $m$.
\paragraph{Quantitative Analysis}
$f_{A}$ and its sub-features have the most positive influence on $m$, which means \emph{acting} is the most important feature for \emph{The Post}. 
%According to \emph{boundedness}, $\left. \nabla \right|_{f'_{A1} \mapsto m} < \left. \nabla \right|_{f_{A} \mapsto m}$ and $\left. \nabla \right|_{f'_{A2} \mapsto m} < \left. \nabla \right|_{f_{A} \mapsto m}$
The attribution of $f_{D}$ is slightly less than for  $f_{A}$ and 
% $\left. \nabla \right|_{f_{D} \mapsto m}$ is slightly less than $\left. \nabla \right|_{f_{A} \mapsto m}$. 
$f_{W}$ has the highest negative influence on $m$. Despite \emph{writing}'s negative influence, \emph{acting} and \emph{directing} still make the movie high-rated.
% \paragraph{Visualisation}
% Visual results of AAEs are shown in Figure \ref{fig_rank_movie}.

\vspace*{-0.2cm}
\paragraph{Property Analysis}
Here, we mainly focus on the \emph{faithfulness} of AAEs. Any disconnected arguments, like $f'_{A2}$ and $f_{D}$, should be assigned attribution scores as $0$, which satisfies \emph{faithfulness} of AAEs. 
From Table \ref{tab_recommender_system}, removing any arguments with positive (negative) influence on $m$ will give rise to a decrease (increase) of the strength of $m$, which guarantees the \emph{faithfulness} of AAEs from the perspective of one single argument.
For example, if $\tau(f_{A})$ is set to $0$, then $\sigma_{f_{A}}'(m)$ will decrease, hence $\sigma_{f_{A}}'(m)-\sigma(m)<0$.
 %any two arguments, like 
 Arguments $f_{A}$ and $f_{D}$ are assigned different contributions, therefore their influence on $m$ is different. 
For instance,  $\left| \tau(f_{A}) \cdot \left. \nabla \right|_{f_{A} \mapsto m} \right|=0.02820 > \left| \tau(f_{D}) \cdot \left. \nabla \right|_{f_{D} \mapsto m} \right|=0.00779$, 
hence $\left| \sigma_{f_{A}}'(m)-\sigma(m) \right|=0.02820 > \left| \sigma_{f_{D}}'(m)-\sigma(m) \right|=0.00779$, which guarantees the \emph{faithfulness} of AAEs from the perspective of comparing %any 
the two arguments.

% \begin{table}[h]
% \caption{Sorted attribution of each argument}
% \label{tab_recommender_system}
% \vskip 0.15in
% \begin{center}
% \begin{small}
% \begin{sc}
% \begin{tabular}{ccc}
% \toprule
% Argument & Attribution & Score \\
% \midrule
% Acting    &$\left. \nabla \right|_{f_{A} \mapsto m}$          &0.176\\
% Actor2    &$\left. \nabla_{\phi} \right|_{f'_{A2} \mapsto m}$ &0.159\\
% Actor1    &$\left. \nabla_{\phi} \right|_{f'_{A1} \mapsto m}$ &0.156\\
% Directing &$\left. \nabla \right|_{f_{D} \mapsto m}$          &0.155\\
% Themes    &$\left. \nabla \right|_{f_{T} \mapsto m}$          &0\\
% Writing   &$\left. \nabla \right|_{f_{W} \mapsto m}$          &-0.21\\
% \bottomrule
% \end{tabular}
% \end{sc}
% \end{small}
% \end{center}
% \vskip -0.1in
% \end{table}

\section{Conclusions and Future Works}
\label{sec_conclusion}
We introduced AAEs as quantitative explanations for acyclic QBAFs equipped with DF-QuAD gradual semantics and showed that they satisfy several desirable properties.
%Similar to 
Inspired by feature attribution methods  \cite{baehrens2010explain}, AAEs 
quantify the contribution of arguments to the final
strength of a topic argument. As our analysis shows,
the scores are faithful in the sense that they 
represent the true effects on the argument 
(e.g. they satisfy \emph{missingness} and \emph{completeness}).
Furthermore, they are computationally efficient as they
can be computed in linear time with respect to the number of arguments in the QBAF.
Finally,
as a first proof of concept, we demonstrated the applicability of AAEs as quantitative explanations in two simple case studies in fake news detection and movie recommender systems, emphasizing  their added value against qualitative explanations such as conversations.

% We believe that AAEs enrich existing argumentative explanations in three aspects. 
% First, we have quantified the attribution scores for arguments in the underlying QBAF, which is a more clear and intuitive to show the explanations.
% Second, we proposed several desirable properties which guarantee the \emph{faithfulness} of AAEs to the underlying QBAF.
% Third, AAEs are computationally efficient in that the attribution explanations can be extracted in linear time w.r.t. the number of arguments in QBAF.
% Second, our method is low time cost because we have given and proved the analytical attribution value in the proposed theory.
% Third, our theory can help compact the argumentation framework by trim the redundant arguments with low attribution scores.

We are planning to extend this work in four directions.
% For one thing, there are still some interesting properties need to be further explored, such as the monotonicity and the robustness;
First, it would be interesting to extend the scope of AAEs from individual influences to the collective influence of a set of arguments on the same topic argument.
Second, we would like to study AAEs for QBAFs equipped with other gradual semantics, like Euler-based \cite{amgoud2018evaluation} and quadratic energy semantics \cite{Potyka18}% because of the intuitiveness and generality of the AAEs idea
.
Although our proposed properties are tailored to AAEs under the DF-QuAD gradual semantics, they also have the potential to be evaluated for other quantitative argumentative explanations due to their generality (e.g. \emph{counterfactuality} and \emph{monotonicity}).
Third, we would like to extend our approach to cyclic QBAFs. 
Let us note that this is not straightforward because the strength values in cyclic QBAFs are defined by an iterative procedure that does not necessarily converge \cite{mossakowski2018modular,potyka_modular_2019}. 
Fourth, we would like to carry out a number of diverse experiments to further improve the human-friendliness of AAEs, which depends on the humans, their expertise and the field of application.
% Fourth, we would like to carry out user studies of our method to further explore the practicality of AAEs.

% In particular, the limit construction in our current
% definition of AAEs is not necessarily well-defined
% and needs to be adapted.

\ack This research was partially funded by the  European Research Council (ERC) under the
European Union’s Horizon 2020 research and innovation programme (grant
agreement No. 101020934, ADIX) and by J.P. Morgan and by the Royal
Academy of Engineering under the Research Chairs and Senior Research
Fellowships scheme.  Any views or opinions expressed herein are solely those of the authors.

\bibliography{ecai}

\newpage
\setcounter{page}{1}
\onecolumn
\appendix
\section*{Supplementary Material for\\``Argument Attribution Explanations in Quantitative Bipolar Argumentation Frameworks''}
% \todo{the content of the proposition may not be same with that of in the previous content. will change everything in the last minutes}
\medskip
\section{Proofs} 
\subsection{Proofs for Section~\ref{sec_theory}} 
\setcounter{proposition}{0}
\setcounter{property}{0}
\begin{proposition}[Direct Qualitative Attribution Influence]
\label{prop_point_qua_proof}
If $B, A \in \mathcal{A}$ are directly connected, then
    \begin{enumerate}
        \item If $(B, A) \in \mathcal{R^{-}}$, then $\left. \nabla \right|_{B \mapsto A} \leq 0$;
        \item If $(B, A) \in \mathcal{R^{+}}$, then $\left. \nabla \right|_{B \mapsto A} \geq 0$.
    \end{enumerate}
\end{proposition}
\begin{proof}
We prove the claim by considering 4 cases.
\begin{enumerate}
\item $(B, A) \in \mathcal{R^{-}} \wedge v_{Aa} \geq v_{As}$.

According to Proposition \ref{prop_point_quan_proof},
$$
\left. \nabla \right|_{B \mapsto A}=-\tau(A) \cdot (1-\left| v_{Ba}-v_{Bs} \right|) \cdot \prod_{\left \{ Z \in \mathcal{A}\setminus{B} \mid (Z,A) \in \mathcal{R^{-}} \right \} }(1-\sigma(Z))
$$

Based on the definition of Df-QuAD, it is easy to check that $\tau(A) \geq 0$, $(1-\tau(Z)) \geq 0$, $\prod_{\left \{ Z \in \mathcal{A}\setminus{B} \mid (Z,A) \in \mathcal{R^{-}} \right \} }(1-\sigma(Z)) \geq 0$, $v_{Ba}=1-\prod_{\left \{ X \in \mathcal{A} \mid (X,B) \in \mathcal{R^{-}} \right \} }(1-\sigma(X)) \geq 0$, $v_{Bs}=1-\prod_{\left \{ Z \in \mathcal{A} \mid (Z,B) \in \mathcal{R^{+}} \right \} }(1-\sigma(Z)) \geq 0$ and $(1-\left| v_{Ba}-v_{Bs} \right|) \geq 0$.
% $$
% \begin{cases}
% & \tau(A) \geq 0;\\ \\
% & (1-\tau(Z)) \geq 0; \\ \\
% & \prod_{\left \{ Z \in \mathcal{A}\setminus{B} \mid (Z,A) \in \mathcal{R^{-}} \right \} }(1-\sigma(Z)) \geq 0;\\ \\
% & v_{Ba}=1-\prod_{\left \{ X \in \mathcal{A} \mid (X,B) \in \mathcal{R^{-}} \right \} }(1-\sigma(X)) \geq 0\\ \\
% & v_{Bs}=1-\prod_{\left \{ Z \in \mathcal{A} \mid (Z,B) \in \mathcal{R^{+}} \right \} }(1-\sigma(Z)) \geq 0\\ \\
% & (1-\left| v_{Ba}-v_{Bs} \right|) \geq 0
% \end{cases}
% $$
Therefore, $\left. \nabla \right|_{B \mapsto A} \leq 0$. The equality holds if and only if $\tau(A)=0$.
\item $(B, A) \in \mathcal{R^{-}} \wedge v_{Aa} < v_{As}$.

According to Proposition \ref{prop_point_quan_proof},
$$
\left. \nabla \right|_{B \mapsto A}=(\tau(A)-1) \cdot (1-\left| v_{Ba}-v_{Bs} \right|) \cdot \prod_{\left \{ Z \in \mathcal{A}\setminus{B} \mid (Z,A) \in \mathcal{R^{-}} \right \} }(1-\sigma(Z))
$$
Similar to before, $(\tau(A)-1) \leq 0$, $(1-\left| v_{Ba}-v_{Bs} \right|) \geq 0$ and $\prod_{\left \{ Z \in \mathcal{A}\setminus{B} \mid (Z,A) \in \mathcal{R^{-}} \right \} }(1-\sigma(Z)) \geq 0$.
Therefore, $\left. \nabla \right|_{B \mapsto A} \leq 0$. The equality holds if and only if $\tau(A)=1$.

\item $(B, A) \in \mathcal{R^{+}} \wedge v_{Aa} \leq v_{As}$.

According to Proposition \ref{prop_point_quan_proof},
$$
\left. \nabla \right|_{B \mapsto A}=\tau(A) \cdot (1-\left| v_{Ba}-v_{Bs} \right|) \cdot \prod_{\left \{ Z \in \mathcal{A}\setminus{B} \mid (Z,A) \in \mathcal{R^{+}} \right \} }(1-\sigma(Z))
$$
Similar to before,  $\tau(A) \geq 0$, $(1-\left| v_{Ba}-v_{Bs} \right|) \geq 0$ and $\prod_{\left \{ Z \in \mathcal{A}\setminus{B} \mid (Z,A) \in \mathcal{R^{+}} \right \} }(1-\sigma(Z)) \geq 0$.
Therefore, $\left. \nabla \right|_{B \mapsto A} \geq 0$. The equality holds if and only if $\tau(A)=0$.

\item $(B, A) \in \mathcal{R^{+}} \wedge v_{Aa} > v_{As}$.

According to Proposition \ref{prop_point_quan_proof},
$$
\left. \nabla \right|_{B \mapsto A}=(1-\tau(A)) \cdot (1-\left| v_{Ba}-v_{Bs} \right|) \cdot \prod_{\left \{ Z \in \mathcal{A}\setminus{B} \mid (Z,A) \in \mathcal{R^{+}} \right \} }(1-\sigma(Z))
$$
Similar to before,  $(1-\tau(A)) \geq 0$, $(1-\left| v_{Ba}-v_{Bs} \right|) \geq 0$ and $\prod_{\left \{ Z \in \mathcal{A}\setminus{B} \mid (Z,A) \in \mathcal{R^{+}} \right \} }(1-\sigma(Z)) \geq 0$.
Therefore, $\left. \nabla \right|_{B \mapsto A} \geq 0$. The equality holds if and only if $\tau(A)=1$.
\end{enumerate}
Summarizing: 
    \begin{enumerate}
        \item If $(B, A) \in \mathcal{R^{-}}$, then $\left. \nabla \right|_{B \mapsto A} \leq 0$;
        \item If $(B, A) \in \mathcal{R^{+}}$, then $\left. \nabla \right|_{B \mapsto A} \geq 0$.
    \end{enumerate}
\end{proof}
\begin{proposition}[Direct Quantitative Attribution Influence]
\label{prop_point_quan_proof}
If $B, A \in \mathcal{A}$
are directly connected and $(B,A) \in \mathcal{R^{*}}$, then
$$
\left. \nabla \right|_{B \mapsto A}=\xi_{B}  (1-\left| v_{Ba}-v_{Bs} \right|)  \prod_{\left \{ Z \in \mathcal{A}\setminus{B} \mid (Z,A) \in \mathcal{R^{*}} \right \} }\left[ 1-\sigma(Z) \right]
$$
where $$
\xi_{B}=    
    \begin{cases}
        -\tau(A) & if \ \ \mathcal{R^{*}}=\mathcal{R^{-}} \wedge v_{Aa} \geq v_{As}\\
        (\tau(A)-1) & if \ \ \mathcal{R^{*}}=\mathcal{R^{-}} \wedge v_{Aa} < v_{As}\\
        \tau(A) & if \ \ \mathcal{R^{*}}=\mathcal{R^{+}} \wedge v_{Aa} > v_{As}\\
        (1-\tau(A)) & if \ \ \mathcal{R^{*}}=\mathcal{R^{+}} \wedge v_{Aa} \leq v_{As}\\
    \end{cases}
$$

\end{proposition}

\begin{proof}
First we compute $\sigma_{\varepsilon}(B)-\sigma(B)$.

% \todo{"First we compute" instead of "Now we compute"?}
If $v_{Ba} \geq v_{Bs}$, then 
$$
\sigma(B)=\tau(B)-\tau(B) \cdot (v_{Ba}-v_{Bs})
$$
$$
\sigma_{\varepsilon}(B)=\tau_{\varepsilon}(B)-\tau_{\varepsilon}(B) \cdot (v_{Ba}-v_{Bs})
$$
\begin{equation}
\label{sigmaB1}
    \begin{aligned}
        \sigma_{\varepsilon}(B)-\sigma(B)
        &= (\tau_{\varepsilon}(B)-\tau(B))-(\tau_{\varepsilon}(B)-\tau(B))\cdot(v_{Ba}-v_{Bs})\\
        &= \left[ (\tau(B)+\varepsilon)-\tau(B)) \right]-\left[ (\tau(B)+\varepsilon)-\tau(B)) \right]\cdot(v_{Ba}-v_{Bs})\\
        &= \varepsilon-\varepsilon \cdot(v_{Ba}-v_{Bs})\\
        &= \varepsilon \cdot \left[ 1-(v_{Ba}-v_{Bs}) \right]
    \end{aligned}
\end{equation}

If $v_{Ba} < v_{Bs}$, then 
$$
\sigma(B)=\tau(B)+(1-\tau(B)) \cdot (v_{Bs}-v_{Ba})
$$
$$
\sigma_{\varepsilon}(B)=\tau_{\varepsilon}(B)+(1-\tau_{\varepsilon}(B)) \cdot (v_{Bs}-v_{Ba})
$$
\begin{equation}
\label{sigmaB2}
    \begin{aligned}
        \sigma_{\varepsilon}(B)-\sigma(B)
        &= (\tau_{\varepsilon}(B)-\tau(B))+\left[ (1-\tau_{\varepsilon}(B))-(1-\tau(B)) \right]\cdot(v_{Bs}-v_{Ba})\\
        &= (\tau_{\varepsilon}(B)-\tau(B))-(\tau_{\varepsilon}(B)-\tau(B)) \cdot(v_{Bs}-v_{Ba})\\
        &= \left[ (\tau(B)+\varepsilon)-\tau(B)) \right]-\left[ (\tau(B)+\varepsilon)-\tau(B)) \right]\cdot(v_{Bs}-v_{Ba})\\
        &= \varepsilon-\varepsilon \cdot(v_{Bs}-v_{Ba})\\
        &= \varepsilon \cdot \left[ 1-(v_{Bs}-v_{Ba}) \right]
    \end{aligned}
\end{equation}

According to \eqref{sigmaB1} and \eqref{sigmaB2}, we have
\begin{equation}
\label{sigmaB3}
    \sigma_{\varepsilon}(B)-\sigma(B)=\varepsilon \cdot (1-\left| v_{Ba}-v_{Bs} \right|)
\end{equation}

Next, we compute $\sigma_{\varepsilon}(A)-\sigma(A)$ by considering 4 cases. We use $v'_{Aa}$ and $v'_{As}$ to denote the aggregation strength of attackers and supporters of $A$ after the change of $\tau(B)$. % \todo{"by considering" instead of "which contains"?}
\begin{enumerate}
\item $(B, A) \in \mathcal{R^{-}} \wedge v_{Aa} \geq v_{As}$.

If $\varepsilon \to 0$, then $v'_{Aa}>v'_{As}$, $v'_{As}=v_{As}$ and $v'_{Aa}>v_{Aa}$. 
$$
\sigma(A)=\tau(A)-\tau(A)\cdot(v_{Aa}-v_{As})
$$
$$
\sigma_{\varepsilon}(A)=\tau(A)-\tau(A)\cdot(v'_{Aa}-v'_{As})
$$
\begin{equation}
\label{sigmaA1}
    \begin{aligned}
        \sigma_{\varepsilon}(A)-\sigma(A)
        &= \left[ \tau(A)-\tau(A)\cdot(v'_{Aa}-v'_{As}) \right]-\left[ \tau(A)-\tau(A)\cdot(v_{Aa}-v_{As}) \right]\\
        &= \tau(A)\cdot \left[ (v_{Aa}-v_{As})-(v'_{Aa}-v'_{As}) \right]\\
        &= \tau(A)\cdot \left[ (v_{Aa}-v'_{Aa})-(v_{As}-v'_{As}) \right]
    \end{aligned}
\end{equation}

Since $v'_{As}=v_{As}$, so 
$$\sigma_{\varepsilon}(A)-\sigma(A)=\tau(A)\cdot (v_{Aa}-v'_{Aa})$$
According to DF-QuAD,
\begin{equation}
\label{sigmaAB1}
    \begin{aligned}
        \sigma_{\varepsilon}(A)-\sigma(A)
        &= \tau(A)\cdot \left[ \left( 1-\prod_{\left \{ Z \in \mathcal{A} \mid (Z,A) \in \mathcal{R^{-}} \right \} }(1-\sigma(Z))\right)-\left( 1-\prod_{\left \{ Z \in \mathcal{A} \mid (Z,A) \in \mathcal{R^{-}} \right \} } (1-\sigma'(Z))\right) \right]\\
        &= \tau(A) \cdot \left[ \prod_{\left \{ Z \in \mathcal{A} \mid (Z,A) \in \mathcal{R^{-}} \right \} } (1-\sigma'(Z))-\prod_{\left \{ Z \in \mathcal{A} \mid (Z,A) \in \mathcal{R^{-}} \right \} } (1-\sigma(Z)) \right]\\
        &= \tau(A) \cdot \left[ (1-\sigma_{\varepsilon}(B)) \cdot \prod_{\left \{ Z \in \mathcal{A}\setminus{B} \mid (Z,A) \in \mathcal{R^{-}} \right \} } (1-\sigma'(Z))-(1-\sigma(B)) \cdot \prod_{\left \{ Z \in \mathcal{A}\setminus{B} \mid (Z,A) \in \mathcal{R^{-}} \right \} } (1-\sigma(Z)) \right]\\
        &= \tau(A) \cdot \left[ (1-\sigma_{\varepsilon}(B))-(1-\sigma(B)) \right] \cdot \prod_{\left \{ Z \in \mathcal{A}\setminus{B} \mid (Z,A) \in \mathcal{R^{-}} \right \} } (1-\sigma(Z))\\
        &= -\tau(A) \cdot (\sigma_{\varepsilon}(B)-\sigma(B)) \cdot \prod_{\left \{ Z \in \mathcal{A}\setminus{B} \mid (Z,A) \in \mathcal{R^{-}} \right \} } (1-\sigma(Z))\\
    \end{aligned}
\end{equation}
According to \eqref{sigmaB3}, 
\begin{equation}
    \sigma_{\varepsilon}(A)-\sigma(A)
    = -\tau(A) \cdot \varepsilon \cdot (1-\left| v_{Ba}-v_{Bs} \right|) \cdot \prod_{\left \{ Z \in \mathcal{A}\setminus{B} \mid (Z,A) \in \mathcal{R^{-}} \right \} } (1-\sigma(Z))
\end{equation}

According to Definition \ref{def_attribution},
\begin{equation}
\label{case1}
    \begin{aligned}
        \left. \nabla \right|_{B \mapsto A}= \lim_{\varepsilon \to 0}\frac{\sigma_{\varepsilon}(A)-\sigma(A)}{\varepsilon}= -\tau(A) \cdot (1-\left| v_{Ba}-v_{Bs} \right|) \cdot \prod_{\left \{ Z \in \mathcal{A}\setminus{B} \mid (Z,A) \in \mathcal{R^{-}} \right \} } (1-\sigma(Z))
    \end{aligned}
\end{equation}

\item $(B, A) \in \mathcal{R^{-}} \wedge v_{Aa} < v_{As}$.

If $\varepsilon \to 0$, then $v'_{Aa}<v'_{As}$, $v'_{As}=v_{As}$ and $v'_{Aa}>v_{Aa}$.

$$
\sigma(A)=\tau(A)+(1-\tau(A))\cdot(v_{As}-v_{Aa})
$$
$$
\sigma_{\varepsilon}(A)=\tau(A)+(1-\tau(A))\cdot(v'_{As}-v'_{Aa})
$$
\begin{equation}
\label{sigmaA2}
    \begin{aligned}
        \sigma_{\varepsilon}(A)-\sigma(A)
        &= \left[ \tau(A)+(1-\tau(A)) \cdot(v'_{As}-v'_{Aa}) \right]-\left[ \tau(A)+(1-\tau(A)) \cdot(v_{As}-v_{Aa}) \right]\\
        &= (1-\tau(A)) \cdot \left[ (v'_{As}-v'_{Aa})-(v_{As}-v_{Aa}) \right]\\
        &= (1-\tau(A)) \cdot \left[ (v_{Aa}-v'_{Aa})-(v_{As}-v'_{As}) \right]\\
    \end{aligned}
\end{equation}
Since $v'_{As}=v_{As}$, so 
\begin{equation}
    \begin{aligned}
        \sigma_{\varepsilon}(A)-\sigma(A)
        &= (1-\tau(A)) \cdot (v_{Aa}-v'_{Aa})\\
    \end{aligned}
\end{equation}
According to DF-QuAD,
\begin{equation}
\label{sigmaAB2}
    \begin{aligned}
        \sigma_{\varepsilon}(A)-\sigma(A)
        &= (1-\tau(A)) \cdot \left[ \left( 1-\prod_{\left \{ Z \in \mathcal{A} \mid (Z,A) \in \mathcal{R^{-}} \right \} }(1-\sigma(Z))\right)-\left( 1-\prod_{\left \{ Z \in \mathcal{A} \mid (Z,A) \in \mathcal{R^{-}} \right \} } (1-\sigma'(Z))\right) \right]\\
        &= (1-\tau(A)) \cdot \left[ \prod_{\left \{ Z \in \mathcal{A} \mid (Z,A) \in \mathcal{R^{-}} \right \} } (1-\sigma'(Z))-\prod_{\left \{ Z \in \mathcal{A} \mid (Z,A) \in \mathcal{R^{-}} \right \} } (1-\sigma(Z)) \right]\\
        &= (1-\tau(A)) \cdot \left[ (1-\sigma_{\varepsilon}(B)) \prod_{\left \{ Z \in \mathcal{A}\setminus{B} \mid (Z,A) \in \mathcal{R^{-}} \right \} } (1-\sigma'(Z))-(1-\sigma(B)) \prod_{\left \{ Z \in \mathcal{A}\setminus{B} \mid (Z,A) \in \mathcal{R^{-}} \right \} } (1-\sigma(Z)) \right]\\
        &= (1-\tau(A)) \cdot \left[ (1-\sigma_{\varepsilon}(B))-(1-\sigma(B)) \right] \cdot \prod_{\left \{ Z \in \mathcal{A}\setminus{B} \mid (Z,A) \in \mathcal{R^{-}} \right \} } (1-\sigma(Z))\\
        &= (\tau(A)-1) \cdot (\sigma_{\varepsilon}(B)-\sigma(B)) \cdot \prod_{\left \{ Z \in \mathcal{A}\setminus{B} \mid (Z,A) \in \mathcal{R^{-}} \right \} } (1-\sigma(Z))\\
    \end{aligned}
\end{equation}
According to \eqref{sigmaB3}, 
\begin{equation}
    \begin{aligned}
        \sigma_{\varepsilon}(A)-\sigma(A)
        &= (\tau(A)-1) \cdot \varepsilon \cdot (1-\left| v_{Ba}-v_{Bs} \right|) \cdot \prod_{\left \{ Z \in \mathcal{A}\setminus{B} \mid (Z,A) \in \mathcal{R^{-}} \right \} } (1-\sigma(Z))\\
    \end{aligned}
\end{equation}
According to Definition \ref{def_attribution},
\begin{equation}
\label{case2}
    \begin{aligned}
        \left. \nabla \right|_{B \mapsto A}= \lim_{\varepsilon \to 0}\frac{\sigma_{\varepsilon}(A)-\sigma(A)}{\varepsilon}= (\tau(A)-1) \cdot (1-\left| v_{Ba}-v_{Bs} \right|) \cdot \prod_{\left \{ Z \in \mathcal{A}\setminus{B} \mid (Z,A) \in \mathcal{R^{-}} \right \} } (1-\sigma(Z))
    \end{aligned}
\end{equation}
% Now we consider the case $(B, A) \in \mathcal{R^{+}}$.
% \todo{"Now we consider the case" instead of "We then prove when"?}
% We can use the conclusion of $\sigma_{\varepsilon}(B)-\sigma(B)$ which is proven above.
% \todo{perhaps move the calculation of $\sigma_{\varepsilon}(B)-\sigma(B)$ before 1 and 2 if you use it in both cases. Instead of "We can use the conclusion of $\sigma_{\varepsilon}(B)-\sigma(B)$" perhaps "We use the fact that $\sigma_{\varepsilon}(B)-\sigma(B) = $(write what it is after the equality sign) as we proved before"}
% Now, We compute $\sigma_{\varepsilon}(A)-\sigma(A)$ which also contains two cases.

\item $(B, A) \in \mathcal{R^{+}} \wedge v_{As} > v_{Aa}$.

If $\varepsilon \to 0$, then $v'_{As}>v'_{Aa}$, $v'_{As}>v_{As}$ and $v'_{Aa}=v_{Aa}$.

% Case 1: $v_{As} > v_{Aa}$ and $\varepsilon \to 0$ which satisfy $v'_{As}>v'_{Aa}$

% $$
% \sigma(A)=\tau(A)-\tau(A)\cdot(v_{Aa}-v_{As})
% $$
% $$
% \sigma_{\varepsilon}(A)=\tau(A)-\tau(A)\cdot(v'_{Aa}-v'_{As})
% $$
According to \eqref{sigmaA1},
\begin{equation}
    \begin{aligned}
        \sigma_{\varepsilon}(A)-\sigma(A)
        &= \tau(A)\cdot \left[ (v_{Aa}-v'_{Aa})-(v_{As}-v'_{As}) \right]\\
    \end{aligned}
\end{equation}
Since $v'_{Aa}=v_{Aa}$, so 
\begin{equation}
    \begin{aligned}
        \sigma_{\varepsilon}(A)-\sigma(A)
        &= \tau(A)\cdot (v'_{As}-v_{As})\\
    \end{aligned}
\end{equation}
According to DF-QuAD,
\begin{equation}
\label{sigmaAB3}
    \begin{aligned}
        \sigma_{\varepsilon}(A)-\sigma(A)
        &= \tau(A)\cdot \left[ \left( 1-\prod_{\left \{ Z \in \mathcal{A} \mid (Z,A) \in \mathcal{R^{+}} \right \} }(1-\sigma'(Z))\right)-\left( 1-\prod_{\left \{ Z \in \mathcal{A} \mid (Z,A) \in \mathcal{R^{+}} \right \} } (1-\sigma(Z))\right) \right]\\
        &= \tau(A) \cdot \left[ \prod_{\left \{ Z \in \mathcal{A} \mid (Z,A) \in \mathcal{R^{+}} \right \} } (1-\sigma(Z))-\prod_{\left \{ Z \in \mathcal{A} \mid (Z,A) \in \mathcal{R^{+}} \right \} } (1-\sigma'(Z)) \right]\\
        &= \tau(A) \cdot \left[ (1-\sigma(B)) \cdot \prod_{\left \{ Z \in \mathcal{A}\setminus{B} \mid (Z,A) \in \mathcal{R^{+}} \right \} } (1-\sigma(Z))-(1-\sigma_{\varepsilon}(B)) \cdot \prod_{\left \{ Z \in \mathcal{A}\setminus{B} \mid (Z,A) \in \mathcal{R^{+}} \right \} } (1-\sigma'(Z)) \right]\\
        &= \tau(A) \cdot \left[ (1-\sigma(B))-(1-\sigma_{\varepsilon}(B)) \right] \cdot \prod_{\left \{ Z \in \mathcal{A}\setminus{B} \mid (Z,A) \in \mathcal{R^{+}} \right \} } (1-\sigma(Z))\\
        &= \tau(A) \cdot (\sigma_{\varepsilon}(B)-\sigma(B)) \cdot \prod_{\left \{ Z \in \mathcal{A}\setminus{B} \mid (Z,A) \in \mathcal{R^{+}} \right \} } (1-\sigma(Z))\\
    \end{aligned}
\end{equation}
According to \eqref{sigmaB3}, 
\begin{equation}
    \begin{aligned}
        \sigma_{\varepsilon}(A)-\sigma(A)
        &= \tau(A) \cdot \varepsilon \cdot (1-\left| v_{Ba}-v_{Bs} \right|) \cdot \prod_{\left \{ Z \in \mathcal{A}\setminus{B} \mid (Z,A) \in \mathcal{R^{+}} \right \} } (1-\sigma(Z))
    \end{aligned}
\end{equation}
According to Definition \ref{def_attribution},
\begin{equation}
\label{case3}
    \begin{aligned}
        \left. \nabla \right|_{B \mapsto A}= \lim_{\varepsilon \to 0}\frac{\sigma_{\varepsilon}(A)-\sigma(A)}{\varepsilon}= \tau(A) \cdot (1-\left| v_{Ba}-v_{Bs} \right|) \cdot \prod_{\left \{ Z \in \mathcal{A}\setminus{B} \mid (Z,A) \in \mathcal{R^{+}} \right \} } (1-\sigma(Z))
    \end{aligned}    
\end{equation}

\item $(B, A) \in \mathcal{R^{+}} \wedge v_{As} \geq v_{Aa}$.

If $\varepsilon \to 0$, then $v'_{As}>v'_{Aa}$, $v'_{As}>v_{As}$ and $v'_{Aa}=v_{Aa}$.
% $$
% \sigma(A)=\tau(A)+(1-\tau(A))\cdot(v_{As}-v_{Aa})
% $$
% $$
% \sigma_{\varepsilon}(A)=\tau(A)+(1-\tau(A))\cdot(v'_{As}-v'_{Aa})
% $$

According to \eqref{sigmaA2},
\begin{equation}
    \begin{aligned}
        \sigma_{\varepsilon}(A)-\sigma(A)
        &= \left[ \tau(A)+(1-\tau(A)) \cdot(v'_{As}-v'_{Aa}) \right]-\left[ \tau(A)+(1-\tau(A)) \cdot(v_{As}-v_{Aa}) \right]\\
        &= (1-\tau(A)) \cdot \left[ (v'_{As}-v'_{Aa})-(v_{As}-v_{Aa}) \right]\\
        &= (1-\tau(A)) \cdot \left[ (v_{Aa}-v'_{Aa})-(v_{As}-v'_{As}) \right]\\
    \end{aligned}
\end{equation}
Since $v'_{Aa}=v_{Aa}$, so 
\begin{equation}
    \begin{aligned}
        \sigma_{\varepsilon}(A)-\sigma(A)
        &= (1-\tau(A)) \cdot (v'_{As}-v_{As})\\
    \end{aligned}
\end{equation}
According to DF-QuAD,
\begin{equation}
\label{sigmaAB4}
    \begin{aligned}
        \sigma_{\varepsilon}(A)-\sigma(A)
        &= (1-\tau(A)) \cdot \left[ \left( 1-\prod_{\left \{ Z \in \mathcal{A} \mid (Z,A) \in \mathcal{R^{+}} \right \} }(1-\sigma'(Z))\right)-\left( 1-\prod_{\left \{ Z \in \mathcal{A} \mid (Z,A) \in \mathcal{R^{+}} \right \} } (1-\sigma(Z))\right) \right]\\
        &= (1-\tau(A)) \cdot \left[ \prod_{\left \{ Z \in \mathcal{A} \mid (Z,A) \in \mathcal{R^{+}} \right \} } (1-\sigma(Z))-\prod_{\left \{ Z \in \mathcal{A} \mid (Z,A) \in \mathcal{R^{+}} \right \} } (1-\sigma'(Z)) \right]\\
        &= (1-\tau(A)) \cdot \left[ (1-\sigma(B)) \prod_{\left \{ Z \in \mathcal{A}\setminus{B} \mid (Z,A) \in \mathcal{R^{+}} \right \} } (1-\sigma(Z))-(1-\sigma_{\varepsilon}(B)) \prod_{\left \{ Z \in \mathcal{A}\setminus{B} \mid (Z,A) \in \mathcal{R^{+}} \right \} } (1-\sigma'(Z)) \right]\\
        &= (1-\tau(A)) \cdot \left[ (1-\sigma(B))-(1-\sigma_{\varepsilon}(B)) \right] \cdot \prod_{\left \{ Z \in \mathcal{A}\setminus{B} \mid (Z,A) \in \mathcal{R^{+}} \right \} } (1-\sigma(Z))\\
        &= (1-\tau(A)) \cdot (\sigma_{\varepsilon}(B)-\sigma(B)) \cdot \prod_{\left \{ Z \in \mathcal{A}\setminus{B} \mid (Z,A) \in \mathcal{R^{+}} \right \} } (1-\sigma(Z))\\
    \end{aligned}
\end{equation}
According to \eqref{sigmaB3},
\begin{equation}
    \begin{aligned}
        \sigma_{\varepsilon}(A)-\sigma(A)
        &= (1-\tau(A)) \cdot \varepsilon \cdot (1-\left| v_{Ba}-v_{Bs} \right|) \cdot \prod_{\left \{ Z \in \mathcal{A}\setminus{B} \mid (Z,A) \in \mathcal{R^{+}} \right \} } (1-\sigma(Z))\\
    \end{aligned}
\end{equation}
According to Definition \ref{def_attribution},
\begin{equation}
\label{case4}
    \begin{aligned}
        \left. \nabla \right|_{B \mapsto A}= \lim_{\varepsilon \to 0}\frac{\sigma_{\varepsilon}(A)-\sigma(A)}{\varepsilon}= (1-\tau(A)) \cdot (1-\left| v_{Ba}-v_{Bs} \right|) \cdot \prod_{\left \{ Z \in \mathcal{A}\setminus{B} \mid (Z,A) \in \mathcal{R^{+}} \right \} } (1-\sigma(Z))
    \end{aligned}
\end{equation}
Combining \eqref{case1}, \eqref{case2}, \eqref{case3} and \eqref{case4}, we conclude that: 
$$
\left. \nabla \right|_{B \mapsto A}=\xi_{B}  (1-\left| v_{Ba}-v_{Bs} \right|)  \prod_{\left \{ Z \in \mathcal{A}\setminus{B} \mid (Z,A) \in \mathcal{R^{*}} \right \} }\left[ 1-\sigma(Z) \right]
$$
where $$
\xi_{B}=    
    \begin{cases}
        -\tau(A) & if \ \ \mathcal{R^{*}}=\mathcal{R^{-}} \wedge v_{Aa} \geq v_{As}\\
        (\tau(A)-1) & if \ \ \mathcal{R^{*}}=\mathcal{R^{-}} \wedge v_{Aa} < v_{As}\\
        \tau(A) & if \ \ \mathcal{R^{*}}=\mathcal{R^{+}} \wedge v_{Aa} > v_{As}\\
        (1-\tau(A)) & if \ \ \mathcal{R^{*}}=\mathcal{R^{+}} \wedge v_{Aa} \leq v_{As}\\
    \end{cases}
$$
which completes the proof.
% $$
%     \begin{aligned}
%         \left. \nabla \right|_{B \mapsto A} 
%         &= \frac{\sigma_{\varepsilon}(A)-\sigma(A)}{\tau_{\varepsilon}(B)-\tau(B)}\\
%         &= \lim_{\varepsilon \to 0}\frac{\sigma_{\varepsilon}(A)-\sigma(A)}{(\tau(B)+\varepsilon)-\tau(B)}\\
%         &= \lim_{\varepsilon \to 0}\frac{\sigma_{\varepsilon}(A)-\sigma(A)}{\varepsilon}\\
%         &= (1-\tau(A)) \cdot (1-\left| v_{Ba}-v_{Bs} \right|) \cdot \prod_{\left \{ Z \in \mathcal{A}\setminus{B} \mid (Z,A) \in \mathcal{R^{+}} \right \} } (1-\sigma(Z))\\
%     \end{aligned}
%     $$
\end{enumerate}  
\end{proof}

% \begin{proof}
% $$
% \begin{aligned}
%     \left. \nabla \right|_{B \to A} 
%     &= \lim_{\varepsilon \to 0} \frac{f_{att}(v_{B}+\varepsilon)-f_{att}(v_{B})}{(v_{B}+\varepsilon)-v_{B}}\\
%     &= \frac{v_{A} \cdot \left[ 1-(v_{B}+\varepsilon) \right] - v_{A} \cdot \left[ 1-v_{B} \right] }{(v_{B}+\varepsilon)-v_{B}}
% \end{aligned}
% $$
% where numerator $v_{A} \cdot \left[ 1-(v_{B}+\varepsilon) \right] - v_{A} \cdot \left[ 1-v_{B} \right] < 0$ and denominator$(v_{B}+\varepsilon)-v_{B} > 0$, therefore proportion $\left. \nabla \right|_{B \to A}<0$
% \end{proof}
% From proposition \ref{prop_qua_influence}, we know that attacker argument has a negative influence on its direct adjacent argument; while supporter argument has a positive influence on its direct adjacent argument.

\begin{proposition}[Indirect Qualitative Attribution Influence]
\label{prop_path_qua_proof}
% Given an acyclic QBAF $\mathcal{Q}=\left\langle\mathcal{A}, \mathcal{R}^{-}, \mathcal{R}^{+}, \tau \right\rangle$, $X_{1},\ldots, X_{n} \in \mathcal{A}$.
% and $\mathcal{R} = \mathcal{R}^{-} \cup \mathcal{R}^{+}$. 
Given $X_{1},\ldots, X_{n} \in \mathcal{A}$ and $\mathcal{R} = \mathcal{R}^{-} \cup \mathcal{R}^{+}$.
% and $\Phi_{X_{1} \mapsto X_{n}}=\{\phi\}$.
Suppose $S=\{(X_{1}, X_{2}), (X_{2}, X_{3}),\ldots, (X_{n-1}, X_{n})\} \subseteq \mathcal{R}$ ($n \geq 3$) and $|S \cap \mathcal{R}^{-}|=\Theta$.
If $X_{1},X_{n}\in \mathcal{A}$ are indirectly connected through path $\phi=X_{1},\ldots, X_{n}$, then
    \begin{enumerate}
        \item If $\Theta$ is odd, then $\left. \nabla_{\phi} \right|_{X_{1} \mapsto X_{n}} \leq 0$;
        \item If $\Theta$ is even, then $\left. \nabla_{\phi} \right|_{X_{1} \mapsto X_{n}} \geq 0$.
    \end{enumerate}
\end{proposition}
\begin{proof}
% $$
% \left. \nabla \right|_{X_{1} \mapsto X_{n}}=\frac{\prod_{i=1}^{n-1}\left. \nabla \right|_{X_{i} \mapsto X_{i+1}}}{\prod_{i=2}^{n-1}(1-\left| v_{X_{i}{a}}-v_{X_{i}{s}} \right|)}
% $$
According to Proposition \ref{prop_path_quan_proof},
$$
\left. \nabla_{\phi} \right|_{X_{1} \mapsto X_{n}}=(1-\left| v_{X_{1}{a}}-v_{X_{1}{s}} \right|) \cdot \prod_{i=1}^{n-1}\frac{\left. \nabla \right|_{X_{i} \mapsto X_{i+1}}}{(1-\left| v_{X_{i}{a}}-v_{X_{i}{s}} \right|)}
$$

It is easy to check that ${(1-\left| v_{X_{i}{a}}-v_{X_{i}{s}} \right|)} \geq 0$. Hence, the sign of $\left. \nabla \right|_{X_{1} \mapsto X_{n}}$ depends only on $\prod_{i=1}^{n-1}\left. \nabla \right|_{X_{i} \mapsto X_{i+1}}$. 

According to proposition \ref{prop_point_qua_proof}, if $(X_{i},X_{i+1}) \in \mathcal{R}^{-}$, then $\left. \nabla \right|_{X_{i} \mapsto X_{i+1}} \leq 0$; if $(X_{i},X_{i+1}) \in \mathcal{R}^{+}$, then $\left. \nabla \right|_{X_{i} \mapsto X_{i+1}} \geq 0$.

So, if $\Theta$ is odd, then $\left. \nabla \right|_{X_{1} \mapsto X_{n}} \leq 0$; if $\Theta$ is even, then $\left. \nabla \right|_{X_{1} \mapsto X_{n}} \geq 0$, which completes the proof.
\end{proof}

\begin{proposition}[Indirect Quantitative Attribution Influence]
\label{prop_path_quan_proof}
Given $X_{1},\ldots, X_{n} \in \mathcal{A}$. If $X_{1},X_{n}\in \mathcal{A}$ are indirectly connected through path $\phi=X_{1},\ldots, X_{n}$, then
% Given an acyclic QBAF $\mathcal{Q}=\left\langle\mathcal{A}, \mathcal{R}^{-}, \mathcal{R}^{+}, \tau \right\rangle$, $X_{1},\ldots, X_{n} \in \mathcal{A}$. %and $\mathcal{R} = \mathcal{R}^{-} \cup \mathcal{R}^{+}$. 
% Let $X_{1}$ and $X_{n}$ be indirectly connected such that $\phi=X_{1},\ldots, X_{n}$ and $\Phi_{X_{1} \mapsto X_{n}}=\{\phi\}$. Then
$$
\left. \nabla_{\phi} \right|_{X_{1} \mapsto X_{n}}=(1-\left| v_{X_{1}{a}}-v_{X_{1}{s}} \right|) \cdot \prod_{i=1}^{n-1}\frac{\left. \nabla \right|_{X_{i} \mapsto X_{i+1}}}{(1-\left| v_{X_{i}{a}}-v_{X_{i}{s}} \right|)}
$$
\end{proposition}
\begin{proof}
According to Proposition \ref{prop_point_quan_proof}, we have
$$
\left. \nabla \right|_{X_{n-1} \mapsto X_{n}}=(1-\left| v_{X_{n-1}a}-v_{X_{n-1}s} \right|) \cdot \left. \delta \right|_{X_{n-1} \mapsto X_{n}}
$$
where
$$
\left. \delta \right|_{X_{n-1} \mapsto X_{n}}= 
    \begin{cases}
        -\tau(X_{n}) \cdot \prod_{\left \{ Z \in \mathcal{A}\setminus{X_{n-1}} \mid (Z,X_{n}) \in \mathcal{R^{-}} \right \} }(1-\sigma(Z)) & if \ \ (X_{n-1}, X_{n}) \in \mathcal{R^{-}} \wedge v_{X_{n}a} \geq v_{X_{n}s}\\ \\
        (\tau(X_{n})-1) \cdot \prod_{\left \{ Z \in \mathcal{A}\setminus{X_{n-1}} \mid (Z,X_{n}) \in \mathcal{R^{-}} \right \} }(1-\sigma(Z)) & if \ \ (X_{n-1}, X_{n}) \in \mathcal{R^{-}} \wedge v_{X_{n}a} < v_{X_{n}s}\\ \\
        \tau(X_{n}) \cdot \prod_{\left \{ Z \in \mathcal{A}\setminus{X_{n-1}} \mid (Z,X_{n}) \in \mathcal{R^{+}} \right \} }(1-\sigma(Z)) & if \ \ (X_{n-1}, X_{n}) \in \mathcal{R^{+}} \wedge v_{X_{n}a} > v_{X_{n}s}\\ \\
        (1-\tau(X_{n})) \cdot \prod_{\left \{ Z \in \mathcal{A}\setminus{X_{n-1}} \mid (Z,X_{n}) \in \mathcal{R^{+}} \right \} }(1-\sigma(Z)) & if \ \ (X_{n-1}, X_{n}) \in \mathcal{R^{+}} \wedge v_{X_{n}a} \leq v_{X_{n}s}        
    \end{cases}
$$
According to \eqref{sigmaAB1}, \eqref{sigmaAB2}, \eqref{sigmaAB3} and \eqref{sigmaAB4}, we have 
$$
\sigma_{\varepsilon}(X_{n})-\sigma(X_{n})=\left[ \sigma_{\varepsilon}(X_{n-1})-\sigma(X_{n-1}) \right] \cdot \left. \delta \right|_{X_{n-1} \mapsto X_{n}}
$$
% Based on the deduction in the proof of proposition \ref{prop_quan_influence}, we have
% $$
% \sigma_{\varepsilon}(X_{n})-\sigma(X_{n})=\left[ \sigma_{\varepsilon}(X_{n-1})-\sigma(X_{n-1}) \right] \cdot \left. \delta \right|_{X_{n-1} \mapsto X_{n}}
% $$
Therefore,
$$
\begin{aligned}
    & \frac{\sigma_{\varepsilon}(X_{n})-\sigma(X_{n})}{\sigma_{\varepsilon}(X_{n-1})-\sigma(X_{n-1})}=\left. \delta \right|_{X_{n-1} \mapsto X_{n}}\\
    & \frac{\sigma_{\varepsilon}(X_{n-1})-\sigma(X_{n-1})}{\sigma_{\varepsilon}(X_{n-2})-\sigma(X_{n-2})}=\left. \delta \right|_{X_{n-2} \mapsto X_{n-1}}\\
    & \ \ \ \ \ \ \ \ \ \ \ \ \vdots \\
    & \frac{\sigma_{\varepsilon}(X_{2})-\sigma(X_{2})}{\sigma_{\varepsilon}(X_{1})-\sigma(X_{1})}=\left. \delta \right|_{X_{1} \mapsto X_{2}}\\
\end{aligned}
$$
By applying these equalities successively, we get
$$
\frac{\sigma_{\varepsilon}(X_{n})-\sigma(X_{n})}{\sigma_{\varepsilon}(X_{1})-\sigma(X_{1})}=\left. \delta \right|_{X_{1} \mapsto X_{2}} \cdot \left. \delta \right|_{X_{2} \mapsto X_{3}} \ldots \left. \delta \right|_{X_{n-1} \mapsto X_{n}}=\prod_{i=1}^{n-1}\left. \delta \right|_{X_{i} \mapsto X_{i+1}}
$$
Thus, 
$$
\begin{aligned}
    \sigma_{\varepsilon}(X_{n})-\sigma(X_{n})
    &= \left[ \sigma_{\varepsilon}(X_{1})-\sigma(X_{1})\right] \cdot \prod_{i=1}^{n-1}\left. \delta \right|_{X_{i} \mapsto X_{i+1}}\\
\end{aligned}
$$
According to \eqref{sigmaB3}, 
$$
\begin{aligned}
    \sigma_{\varepsilon}(X_{n})-\sigma(X_{n})
    &=\varepsilon \cdot (1-\left| v_{X_{1}}{a}-v_{X_{1}}{s} \right|) \cdot \prod_{i=1}^{n-1}\left. \delta \right|_{X_{i} \mapsto X_{i+1}}
\end{aligned}
$$
According to Definition \ref{def_attribution},
$$
\begin{aligned}
    \left. \nabla \right|_{X_{1} \mapsto X_{n}}
    &= \lim_{\varepsilon \to 0}\frac{\sigma_{\varepsilon}(X_{n})-\sigma(X_{n})}{\varepsilon}\\
    &= (1-\left| v_{X_{1}}{a}-v_{X_{1}}{s} \right|) \cdot \prod_{i=1}^{n-1}\left. \delta \right|_{X_{i} \mapsto X_{i+1}}\\
    &= (1-\left| v_{X_{1}{a}}-v_{X_{1}{s}} \right|) \cdot \prod_{i=1}^{n-1}\frac{\left. \nabla \right|_{X_{i} \mapsto X_{i+1}}}{(1-\left| v_{X_{i}{a}}-v_{X_{i}{s}} \right|)}
\end{aligned}
$$
which completes the proof.
\end{proof}

\subsection{Proofs for Section~\ref{sec_properties}} 
\begin{proposition}[Explainability]
\label{proposition_explainability_proof}
% Let $\mathcal{Q}=\left\langle\mathcal{A}, \mathcal{R}^{-}, \mathcal{R}^{+}, \tau \right\rangle$ be an acyclic QBAF. 
$\forall A,B \in \mathcal{A}$,
$
\left. \nabla \right|_{B \mapsto A} \in \mathbb{R}
$ is well-defined.
\end{proposition}
\begin{proof}
The claim follows by observing that the DF-QuAD gradual semantics $\sigma$  can be written in
closed-form (apply the function recursively to 
the predecessors of arguments - since the graph
is acyclic, this procedure will end after a finite
number of steps) and is differentiable (because it is
composed of differentiable functions), which means
that the limit in the definition of AAE exists.
\end{proof}

\begin{proposition}[Missingness]
\label{proposition_missingness_proof}
% Let $\mathcal{Q}=\left\langle\mathcal{A}, \mathcal{R}^{-}, \mathcal{R}^{+}, \tau \right\rangle$ be an acyclic QBAF.
$\forall A, B \in \mathcal{A}$, if $B$ is not connected to $A$, then
$$
\left. \nabla \right|_{B \mapsto A} = 0.
$$
\end{proposition}
\begin{proof}
If $B$ is not connected to $A$, then $\left| \Phi_{B \mapsto A} \right|=0$ and $\sigma(A)$
does not depend on $\tau(B)$.
% , which denotes there is no path reaching from $B$ to $A$. 
Therefore, $\sigma_{\varepsilon}(A)=\sigma(A)$. 
Then, according to Definition \ref{def_attribution}, 
\begin{equation}
    \begin{aligned}
        \left. \nabla \right|_{B \mapsto A}= \lim_{\varepsilon \to 0}\frac{\sigma_{\varepsilon}(A)-\sigma(A)}{\varepsilon}=0
    \end{aligned}
\end{equation}
\end{proof}

\begin{property}[Completeness]
\label{property_completeness_proof}
Let $A, B \in \mathcal{A}$ and let $\sigma'_B(A)$ denote the strength of $A$ when setting $\tau(B)$ to $0$. Then 
$$
   -\tau(B) \cdot \left. \nabla \right|_{B \mapsto A} = \sigma_{B}'(A)-\sigma(A)
$$
\end{property}

\begin{proposition}
\label{proposition_completeness_satisfy_proof}
Completeness is satisfied if $B$ is directly or indirectly connected to $A$.
\end{proposition}
\begin{proof}
1. Direct connectivity:

According to later Proposition \ref{proposition_quan_invar_satisfy_proof}, 
there exists a constant $C\in [-1,1]$ such that $\left. \nabla \right|_{B \mapsto A} \equiv C$ holds $\forall \tau(B) \in [0,1]$.
Therefore, we have
$$
\left. \nabla \right|_{B \mapsto A} = \frac{\sigma_{B}'(A)-\sigma(A)}{0-\tau(B)},
$$

hence,
$$
\sigma_{B}'(A)-\sigma(A) = -\tau(B) \cdot \left. \nabla \right|_{B \mapsto A} 
$$

2. Indirect connectivity:

According to later Proposition \ref{proposition_quan_invar_satisfy_proof},
there exists a constant $C\in [-1,1]$ such that $\left. \nabla%_{\phi}
\right|_{B \mapsto A} \equiv C$ holds $\forall \tau(B) \in [0,1]$.
Therefore, we have
$$
\left. \nabla%_{\phi}
\right|_{B \mapsto A} = \frac{\sigma_{B}'(A)-\sigma(A)}{0-\tau(B)},
$$

hence,
$$
\sigma_{B}'(A)-\sigma(A) = -\tau(B) \cdot \left. \nabla%_{\phi}
\right|_{B \mapsto A}, 
$$
which completes the proof.
\end{proof}

\begin{proposition}
\label{proposition_completeness_violate_proof}
Completeness can be violated if $B$ is multifold connected to $A$.
\end{proposition}
\begin{proof}
A counter example is shown in Figure~\ref{fig_counter1}.
Suppose the base score $\tau$ of every argument is 0.5 in this QBAF. 
According to DF-QuAD, we have $\sigma(A)=0.96875$, $\sigma(B)=0.75$, $\sigma(C)=0.75$, $\sigma(D)=0.5$. 
According to Definition~\ref{def_attribution}, we have $\left. \nabla \right|_{D \mapsto A}=0.125$.
When setting $\tau(D)$ to $0$, we have $\sigma_{D}'(A)=0.875$.
Therefore, 
$$
\sigma_{D}'(A)-\sigma(A) =-0.09375 \neq
-\tau(D) \cdot \left. \nabla \right|_{D \mapsto A}=-0.0625
$$
which violates completeness.
\begin{figure}[ht]
\centering
\includegraphics[width=0.25\columnwidth]{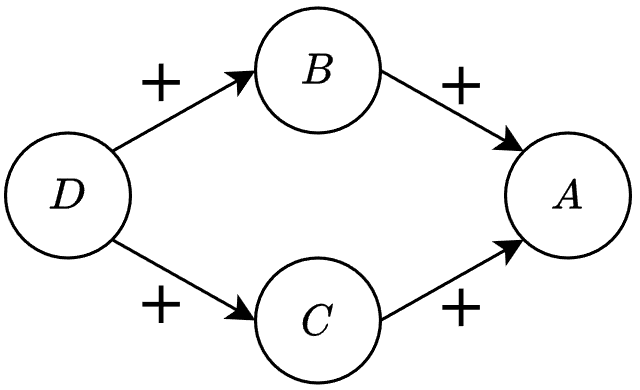}
\caption{A counter example.}
\label{fig_counter1}
\end{figure}
\end{proof}

\begin{property}[Counterfactuality]
\label{property_counterfactuality_proof}
Let $A, B \in \mathcal{A}$ and let $\sigma'_B(A)$ denote the strength of $A$ when setting $\tau(B)$ to $0$. Then
\begin{enumerate}
    \item If $\left. \nabla \right|_{B \mapsto A} \leq 0$, then $\sigma_{B}'(A) \geq \sigma(A)$;
    \item If $\left. \nabla \right|_{B \mapsto A} \geq 0$, then $\sigma_{B}'(A) \leq \sigma(A)$.
\end{enumerate}
\end{property}

\begin{proposition}
\label{proposition_counterfactuality_satisfy_proof}
Counterfactuality is satisfied if $B$ is directly or indirectly connected to $A$.
\end{proposition}
\begin{proof}
%WHY DOES IT PLAY A ROLE HERE? According to Proposition \ref{proposition_completeness_satisfy_proof},
%
if $\left. \nabla \right|_{B \mapsto A} \leq 0$, then $-\tau(B) \cdot \left. \nabla%_{\phi}
\right|_{B \mapsto A} \geq 0$. Thus, 
according to Proposition \ref{proposition_completeness_satisfy_proof}, $\sigma_{B}'(A) \geq \sigma(A)$.
Similarly, if $\left. \nabla \right|_{B \mapsto A} \geq 0$, then $-\tau(B) \cdot \left. \nabla\right|_{B \mapsto A} \leq 0$. Thus, $\sigma_{B}'(A) \leq \sigma(A)$.
\end{proof} 

\begin{proposition}
\label{proposition_counterfactuality_violate_proof}
Counterfactuality can be violated if $B$ is multifold connected to $A$.
\end{proposition}
\begin{proof}
A counter example is shown in Figure~\ref{fig_counter3}, 
where $E$ is multifold connected to $A$. We set $\tau(A)=\tau(B)=\tau(C)=\tau(D)=0$, and $\tau(E)=0.6$.
According to DF-QuAD, we have $\sigma(A)=0.316$.
According to Definition~\ref{def_attribution}, we have $\left. \nabla \right|_{E \mapsto A}=-0.18$.
If we set $\tau(E)$ to $0$, then we have $\sigma_{E}'(A)=0.1<\sigma(A)=0.316$, which violates counterfactuality.

\begin{figure}[ht]
\centering
\includegraphics[width=0.25\columnwidth]{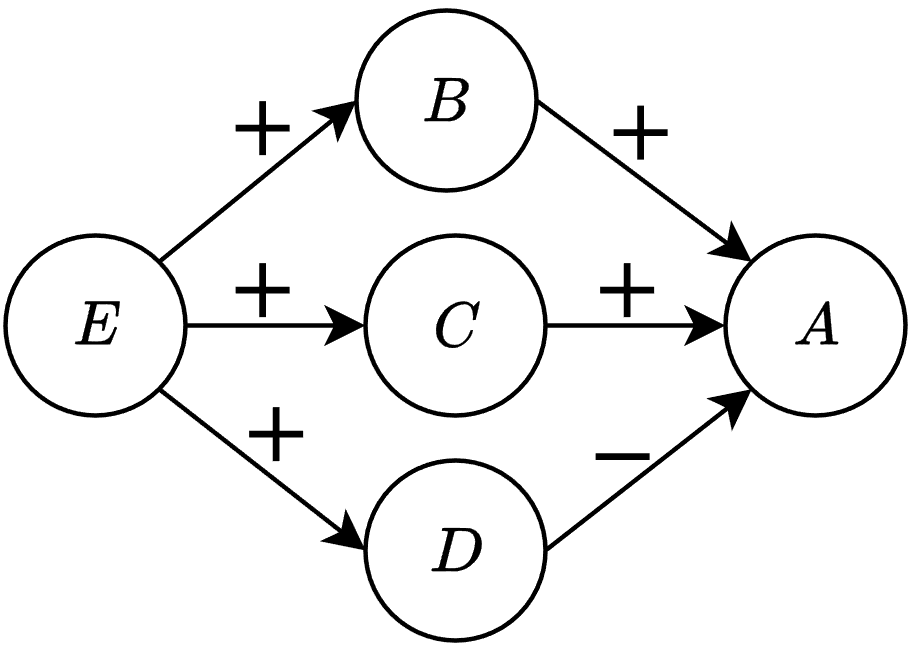}
\caption{A counter example.}
\label{fig_counter3}
\end{figure}

\end{proof}

\begin{property}[Agreement]
\label{property_agreement_proof}
Let $A, B, C \in \mathcal{A}$ and $\sigma'_B(A),\sigma'_C(A)$ denote the strength of $A$ when setting $\tau(B),\tau(C)$ to $0$ respectively.\\
If $$\left| \tau(B) \cdot \left. \nabla \right|_{B \mapsto A} \right| = \left| \tau(C) \cdot \left. \nabla \right|_{C \mapsto A} \right|$$
then $$\left| \sigma_{B}'(A)-\sigma(A) \right| = \left| \sigma_{C}'(A)-\sigma(A) \right|.$$
\end{property}

\begin{proposition}
\label{proposition_agreement_satisfy_proof}
Agreement is satisfied if $B$ and $C$ are directly or indirectly connected to $A$.
\end{proposition}
\begin{proof}
According to Proposition \ref{proposition_completeness_satisfy_proof},
$\left| \tau(B) \cdot \left. \nabla \right|_{B \mapsto A} \right|=\left| \sigma_{B}'(A)-\sigma(A) \right|$ and
$\left| \tau(C) \cdot \left. \nabla \right|_{C \mapsto A} \right|=\left| \sigma_{C}'(A)-\sigma(A) \right|$.
Therefore, if $\left| \tau(B) \cdot \left. \nabla \right|_{B \mapsto A} \right| = \left| \tau(C) \cdot \left. \nabla \right|_{C \mapsto A} \right|$,
then $\left| \sigma_{B}'(A)-\sigma(A) \right| = \left| \sigma_{C}'(A)-\sigma(A) \right|$, 
which completes the proof.
\end{proof}

\begin{proposition}
\label{proposition_agreement_violate_proof}
Agreement can be violated if $B$ or $C$ is multifold connected to $A$.
\end{proposition}
\begin{proof}
A counter example is shown in Figure~\ref{fig_counter2}, 
where $D$ is multifold connected to $A$, and $E$ is directly connected to $A$. We set $\tau(A)=\tau(B)=\tau(C)=\tau(D)=0.5$, and $\tau(E)=0.125$.
According to DF-QuAD, we have $\sigma(A)=0.90625$.
According to Definition~\ref{def_attribution}, we have $\left. \nabla \right|_{D \mapsto A}=0.125$ and $\left. \nabla \right|_{E \mapsto A}=-0.5$.
If we set $\tau(D)$ to $0$, then we have $\sigma_{D}'(A)=0.8125$; if we set $\tau(E)$ to $0$, then we have $\sigma_{E}'(A)=0.96875$.
$\left| \tau(D) \cdot \left. \nabla \right|_{D \mapsto A} \right|=\left| \tau(E) \cdot \left. \nabla \right|_{E \mapsto A} \right|=0.0625$, however, $\left| \sigma_{D}'(A)-\sigma(A) \right|=0.09375 \neq \left| \sigma_{E}'(A)-\sigma(A) \right|=0.0625$,
which violates agreement.
\begin{figure}[ht]
\centering
\includegraphics[width=0.35\columnwidth]{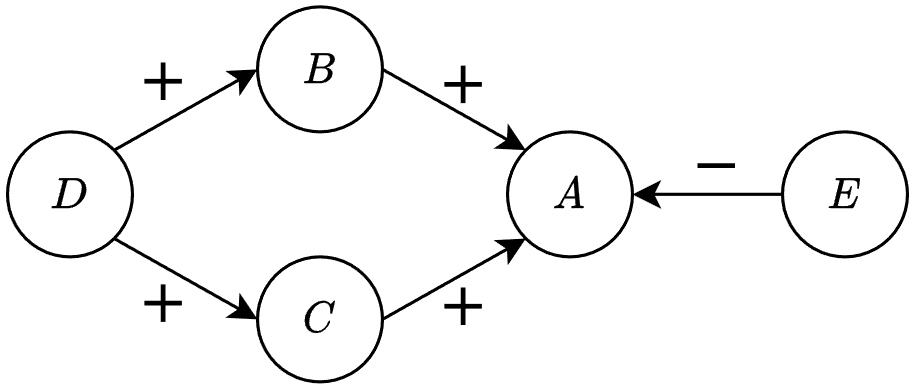}
\caption{A counter example.}
\label{fig_counter2}
\end{figure}
\end{proof}

\begin{property}[Monotonicity]
\label{property_monotonicity_proof}
Let $A, B, C \in \mathcal{A}$, $\sigma'_B(A),\sigma'_C(A)$ denote the strength of $A$ when setting $\tau(B),\tau(C)$ to $0$ respectively.\\
If $$\left| \tau(B) \cdot \left. \nabla \right|_{B \mapsto A} \right| < \left| \tau(C) \cdot \left. \nabla \right|_{C \mapsto A} \right|$$ 
then $$\left| \sigma_{B}'(A)-\sigma(A) \right| < \left| \sigma_{C}'(A)-\sigma(A) \right|.$$
\end{property}

\begin{proposition}
\label{proposition_monotonicity_satisfy_proof}
Monotonicity is satisfied if $B$ and $C$ are directly or indirectly connected to $A$.
\end{proposition}
\begin{proof}
According to Proposition \ref{proposition_completeness_satisfy_proof},
$\left| \tau(B) \cdot \left. \nabla \right|_{B \mapsto A} \right|=\left| \sigma_{B}'(A)-\sigma(A) \right|$ and
$\left| \tau(C) \cdot \left. \nabla \right|_{C \mapsto A} \right|=\left| \sigma_{C}'(A)-\sigma(A) \right|$.
Therefore, if $\left| \tau(B) \cdot \left. \nabla \right|_{B \mapsto A} \right| < \left| \tau(C) \cdot \left. \nabla \right|_{C \mapsto A} \right|$,
then $\left| \sigma_{B}'(A)-\sigma(A) \right| < \left| \sigma_{C}'(A)-\sigma(A) \right|$, 
which completes the proof.
\end{proof}

\begin{proposition}
\label{proposition_monotonicity_violate_proof}
Monotonicity can be violated if $B$ or $C$ is multifold connected to $A$.
\end{proposition}
\begin{proof}
A counter example is shown in Figure~\ref{fig_counter2}, 
where $D$ is multifold connected to $A$, and $E$ is directly connected to $A$. We set $\tau(A)=\tau(B)=\tau(C)=\tau(D)=0.5$, and $\tau(E)=0.15$.
According to DF-QuAD, we have $\sigma(A)=0.89375$.
According to Definition~\ref{def_attribution}, we have $\left. \nabla \right|_{D \mapsto A}=0.125$ and $\left. \nabla \right|_{E \mapsto A}=-0.5$.
If we set $\tau(D)$ to $0$, then we have $\sigma_{D}'(A)=0.8$; if we set $\tau(E)$ to $0$, then we have $\sigma_{E}'(A)=0.96875$.
$\left| \tau(D) \cdot \left. \nabla \right|_{D \mapsto A} \right|=0.0625<\left| \tau(E) \cdot \left. \nabla \right|_{E \mapsto A} \right|=0.075$, however, $\left| \sigma_{D}'(A)-\sigma(A) \right|=0.09375 > \left| \sigma_{E}'(A)-\sigma(A) \right|=0.075$,
which violates monotonicity.
\end{proof}

\begin{property}[Qualitative Invariability]
\label{property_qua_invar_proof}
$\forall A, B \in \mathcal{A}$, let $\nabla_{\delta}$ denote AAE from $B$ to $A$ when setting $\tau(B)$ to some $\delta \in [0,1]$. Then
\begin{enumerate}
    \item If $\left. \nabla \right|_{B \mapsto A} \leq 0$, then $\forall \delta \in [0,1]$, $\nabla_{\delta} \leq 0$;
    \item If $\left. \nabla \right|_{B \mapsto A} \geq 0$, then $\forall \delta \in [0,1]$, $\nabla_{\delta} \geq 0$.
\end{enumerate}
\end{property}

\begin{proposition}
\label{proposition_qua_invar_satisfy_proof}
Qualitative invariability is satisfied if $B$ is directly or indirectly connected to $A$.
\end{proposition}
\begin{proof}
According to later Proposition~\ref{proposition_quan_invar_satisfy_proof}, if $C \leq 0$, then $\nabla_{\delta} \leq 0$ holds $\forall \delta \in [0,1]$; if $C \geq 0$, then $\nabla_{\delta} \geq 0$ holds $\forall \delta \in [0,1]$.
\end{proof}

\begin{proposition}
\label{proposition_qua_invar_violate_proof}
Qualitative invariability can be violated if $B$ is multifold connected to $A$.
\end{proposition}
\begin{proof}
A counter example is shown in Figure~\ref{fig_counter3}, 
where $E$ is multifold connected to $A$. We set $\tau(A)=0.1$, $\tau(B)=\tau(C)=\tau(D)=0$, and $\tau(E)=0.6$.
% According to DF-QuAD, we have $\sigma(A)=0.316$.
According to Definition~\ref{def_attribution}, we have $\left. \nabla \right|_{E \mapsto A}=-0.18<0$.
We let $\delta=0.4$, then we have $\left. \nabla \right|_{E \mapsto A}=0.18>0$, which violates qualitative invariability.
\end{proof}

\begin{property}[Quantitative Invariability]
\label{property_quan_invar_proof}
$\forall A, B \in \mathcal{A}$, %$C \in \mathbb{R}$, 
let $\nabla_{\delta}$ denote the AAE from $B$ to $A$ when setting $\tau(B)$ to some $\delta \in [0,1]$.
For all $\delta$, there always exists a constant $C \in [-1,1]$ such that
$$
\nabla_{\delta} \equiv C
$$
\end{property}

\begin{proposition}
\label{proposition_quan_invar_satisfy_proof}
Quantitative invariability is satisfied if $B$ is directly or indirectly connected to $A$.
\end{proposition}
\begin{proof}
1. Direct connectivity:

According to Proposition \ref{prop_point_quan_proof},
$$
\left. \nabla \right|_{B \mapsto A}=\xi_{B}  (1-\left| v_{Ba}-v_{Bs} \right|)  \prod_{\left \{ Z \in \mathcal{A}\setminus{B} \mid (Z,A) \in \mathcal{R^{*}} \right \} }\left[ 1-\sigma(Z) \right].
$$

Thus, it is easy to check that $\left. \nabla \right|_{B \mapsto A}$ does not depend on $\tau(B)$.
Therefore, $\nabla_{\delta} \equiv C$ holds $\forall \delta \in [0,1]$.\\
2. Indirect connectivity: 

Similarly, based on Proposition \ref{prop_path_quan_proof},
one can check that
$\left. \nabla%_{\phi} 
\right|_{B \mapsto A}$ does not depend on $\tau(B)$. Therefore, $\nabla_{\delta} \equiv C$ holds $\forall \delta \in [0,1]$.
\end{proof}

\begin{proposition}
\label{proposition_quan_invar_violate_proof}
Quantitative invariability can be violated if $B$ is multifold connected to $A$.
\end{proposition}
\begin{proof}
A counter example is shown in Figure~\ref{fig_counter3}, 
where $E$ is multifold connected to $A$. We set $\tau(A)=0.1$, $\tau(B)=\tau(C)=\tau(D)=0$, and $\tau(E)=0.6$.
According to Definition~\ref{def_attribution}, we have $\left. \nabla \right|_{E \mapsto A}=-0.18$.
We let $\delta=0.4$, then we have $\nabla_{\delta}=0.18$. Therefore, there does not exist a constant $C$, which violates quantitative invariability.
\end{proof}

\begin{proposition}[Tractability]
\label{property_tractability}
If $\left| \mathcal{A} \right|=n$, then AAEs can be generated in linear time $\mathcal{O}(n)$.
\end{proposition}
\begin{proof}
To compute the AAE, we first compute a topological ordering of the arguments (linear time). The AAE
is the partial derivative of the strength of
the topic argument with respect to the base score
of the influence argument and can be computed by
the backward propagation procedure used in 
training multilayer perceptrons in linear time \cite{dlbook2016} (instead of inverting the direction from the input neurons to the output neurons,
we invert the direction given by the topological ordering).
\end{proof}

\section{Examples}
% \todo{add some text and motivation before the examples (what does the example illustrate?)}
%In this section
Here, we give more examples %and more details 
to help understand AAEs, the DF-QuAD semantics and 
the notions of direct/indirect attribution influence%in the paper
. %Also, we show an example of applying AAEs for a large QBAF.

\begin{example}[%Attribution Influence
AAE]
\label{example_attribution_influence_appendix}
Consider the acyclic QBAF in Example~\ref{example_DF_QuAD}.
%$\mathcal{Q}=\left\langle\mathcal{A}, \mathcal{R}^{-}, \mathcal{R}^{+}, \tau \right\rangle$ with arguments and relations as in Figure \ref{fig_bcda_two_path} and $\tau(X)=0.5$ for all $X \in \mathcal{A}$.
%\begin{figure}[h]
%	\centering		\includegraphics[width=0.3\columnwidth]{figures/bcda_two_path.png}
%	\caption{An example QBAF structure in Example~\ref{example_attribution_influence_appendix}.}
%	 \label{fig_bcda_two_path_appendix}
%\end{figure}
Using DF-QuAD, we have 
$\sigma(A)=0.375$, $\sigma(B)=0.5$, $\sigma(C)=0.25$, $\sigma(D)=0.375$, $\sigma(F)=0.75$, $\sigma(G)=0.125$.
Then, by Definition~\ref{def_attribution}, we have
$\left. \nabla \right|_{B \mapsto A}=-0.25$, 
$\left. \nabla \right|_{C \mapsto A}=0.125$, 
$\left. \nabla \right|_{D \mapsto A}=-0.375$, 
$\left. \nabla \right|_{F \mapsto A}=-0.125$, 
$\left. \nabla \right|_{G \mapsto A}=0.125$.

\end{example}

\begin{example}[DF-QuAD Semantics]
\label{example_QBAF_small}
Consider an acyclic QBAF $\mathcal{Q}=\left\langle\mathcal{A}, \mathcal{R}^{-}, \mathcal{R}^{+}, \tau \right\rangle$ %equipped with DF-QuAD gradual semantics $\sigma$ 
as in Figure \ref{fig_QBAF_Example_small}, 
assuming that the base score $\tau$ of every argument is 0.5% in $\mathcal{Q}$
.
\begin{figure}[ht]
	\centering
		\includegraphics[width=0.13\columnwidth]{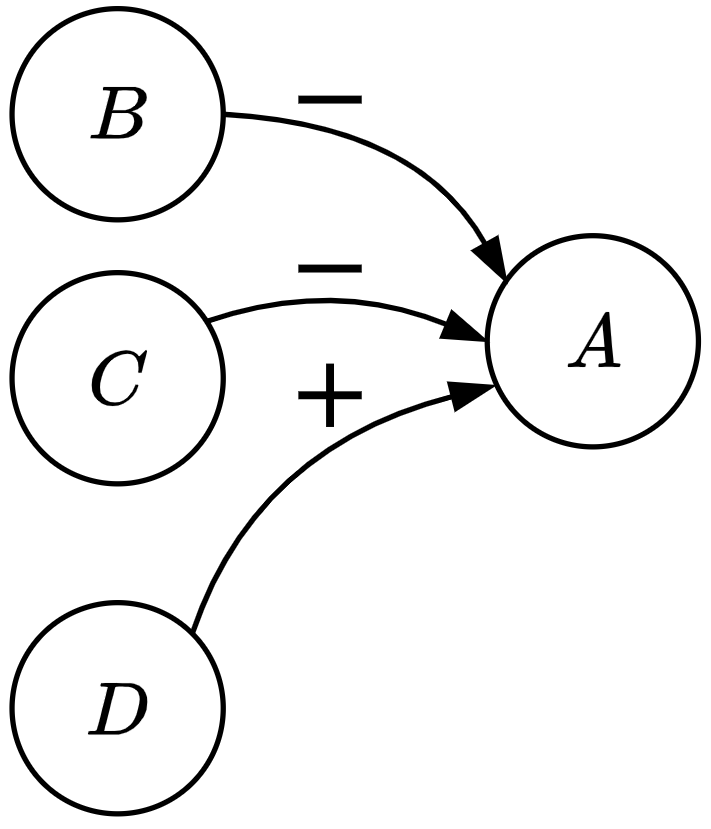}
	\caption{The QBAF in Example \ref{example_QBAF_small}.}
	\label{fig_QBAF_Example_small}
\end{figure}

Arguments $B$, $C$ and $D$ have no supporter and no attacker, so $\sigma(B)=\tau(B)$,  $\sigma(C)=\tau(C)$ and $\sigma(D)=\tau(D)$.

$$
v_{Aa}=1-(1-\sigma(B))(1-\sigma(C))=0.75,
$$
$$
v_{As}=1-(1-\sigma(D))=0.5,
$$
$$
\sigma(A)=\tau(A)-\tau(A)\cdot(v_{Aa}-v_{As})=0.375.
$$
\end{example}

\begin{example}[Direct Attribution Influence]
\label{example_QBAF_abc_appendix}
Consider %an acyclic QBAF $\mathcal{Q}=\left\langle\mathcal{A}, \mathcal{R}^{-}, \mathcal{R}^{+}, \tau \right\rangle$ equipped with DF-QuAD gradual semantics $\sigma$ in Figure \ref{fig_QBAF_Example_abc}. Suppose the base score $\tau$ of every argument is 0.5 in $\mathcal{Q}$.
the QBAF from Example~\ref{example_QBAF_small}.
Let us first consider direct qualitative influence.
According to Proposition \ref{prop_point_qua}, the qualitative influence $\left. \nabla \right|_{B \mapsto A}<0$ because $(B, A) \in \mathcal{R^{-}}$, and the qualitative influence $\left. \nabla \right|_{B \mapsto A}>0$ because $(C, A) \in \mathcal{R^{+}}$.

Next, let us consider direct quantitative influence. 
We have $v_{Aa}=0.5$, $v_{As}=0.5$, $v_{Ba}=v_{Bs}=v_{Ca}=v_{Cs}=0$ and $v_{Aa} \geq v_{As}$.
According to Proposition \ref{prop_point_quan}, 
the quantitative influence
$$
\begin{aligned}
    \left. \nabla \right|_{B \mapsto A}
    &= -\tau(A) \cdot (1-\left| v_{Ba}-v_{Bs} \right|)
    \cdot \prod_{\left \{ Z \in \mathcal{A}\setminus{B} \mid (Z,A) \in \mathcal{R^{-}} \right \} }(1-\sigma(Z))=-0.5,\\
    \left. \nabla \right|_{C \mapsto A}
    &= (1-\tau(A)) \cdot (1-\left| v_{Ca}-v_{Cs} \right|)
    \cdot \prod_{\left \{ Z \in \mathcal{A}\setminus{C} \mid (Z,A) \in \mathcal{R^{+}} \right \} }(1-\sigma(Z))=0.5.\\
\end{aligned}
$$
\end{example}
\begin{figure}[ht]
	\centering
		\includegraphics[width=0.15\columnwidth]{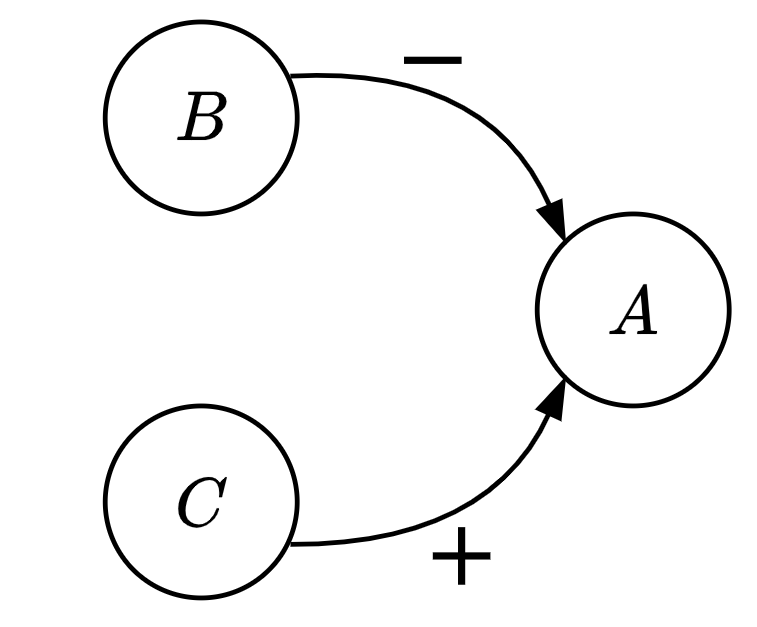}
	\caption{Acyclic QBAF in Example~\ref{example_QBAF_abc_appendix}.}
	\label{fig_QBAF_Example_abc}
\end{figure}
\begin{example}[Indirect Attribution Influence]
\label{example_QBAF}
Consider an acyclic QBAF $\mathcal{Q}=\left\langle\mathcal{A}, \mathcal{R}^{-}, \mathcal{R}^{+}, \tau \right\rangle$ %equipped with DF-QuAD gradual semantics $\sigma$ 
as
in Figure \ref{fig_QBAF_Example}%. Suppose the 
, assuming that the base score $\tau$ of every argument is 0.5% in $\mathcal{Q}$
.\\

\begin{figure}[ht]
	\centering
		\includegraphics[width=0.3\columnwidth]{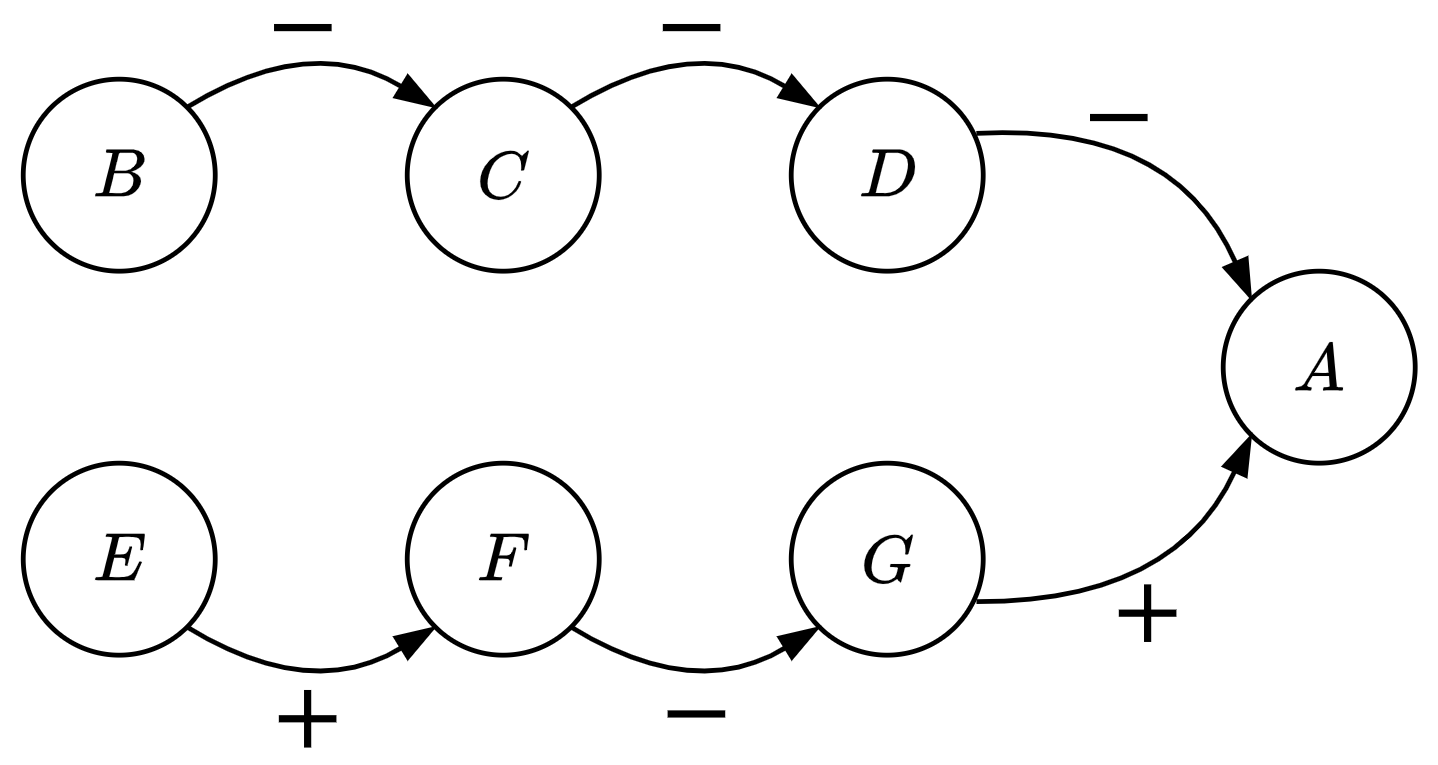}
	\caption{Acyclic QBAF in Example \ref{example_QBAF}.}
	\label{fig_QBAF_Example}
\end{figure}

First, let us consider indirect qualitative influence in Proposition \ref{prop_path_qua}.
For argument $B$ and $D$, we have $S_{B \mapsto D}=\left\{(B,C),(C,D) \right\} \subset \mathcal{R^{-}} \cup \mathcal{R^{+}}$ and $\Theta=\left| S_{B \mapsto D} \cap \mathcal{R^{-}} \right|=2$. The number of %the attack relation 
attackers
is even, so the indirect qualitative influence from argument $B$ to argument $D$ is positive, thus $\left. \nabla \right|_{B \mapsto D}>0$.\\
For argument $E$ and $G$, we have $S_{E \mapsto G}=\left\{(E,F),(F,G) \right\} \subset \mathcal{R^{-}} \cup \mathcal{R^{+}}$ and $\Theta=\left| S_{E \mapsto G} \cap \mathcal{R^{-}} \right|=1$. The number of %the attack relation 
attackers 
is odd, so the indirect qualitative influence from argument $E$ to argument $G$ is negative, thus $\left. \nabla \right|_{E \mapsto G}<0$.

Second, let us consider indirect qualitative influence in Proposition \ref{prop_path_quan}.\\
For argument $B$ and $D$, we have $S_{B \mapsto D}=\left\{(B,C),(C,D) \right\} \subset \mathcal{R^{-}} \cup \mathcal{R^{+}}$, so the indirect quantitative influence from argument $B$ to argument $D$ is 
$$
\begin{aligned}
    \left. \nabla_{\phi} \right|_{B \mapsto D}
    &= \left. \nabla \right|_{B \mapsto C} \times \frac{\left. \nabla \right|_{C \mapsto D}}{1-\left| v_{Ca}-v_{Cs} \right|}\\
    &= -\tau(C) \cdot \prod_{\left \{ Z \in \mathcal{A}\setminus{B} \mid (Z,C) \in \mathcal{R^{-}} \right \} }(1-\sigma(Z))
    \times -\tau(D) \cdot \prod_{\left \{ Z \in \mathcal{A}\setminus{C} \mid (Z,D) \in \mathcal{R^{-}} \right \} }(1-\sigma(Z))\\
    &= \left[-0.5 \times (1-\left| 0-0 \right|) \right] \times (-0.5)\\
    &= 0.25.
\end{aligned}
$$
For argument $E$ and $G$, we have $\phi_{E \mapsto G}=\left\{(E,F),(F,G) \right\} \subset \mathcal{R^{-}} \cup \mathcal{R^{+}}$, so the indirect quantitative influence from argument $E$ to argument $G$ is $$
\begin{aligned}
    \left. \nabla_{\phi} \right|_{E \mapsto G}
    &= \left. \nabla \right|_{E \mapsto F} \times \frac{\left. \nabla \right|_{F \mapsto G}}{1-\left| v_{Fa}-v_{Fs} \right|}\\
    &= -\tau(F) \cdot \prod_{\left \{ Z \in \mathcal{A}\setminus{E} \mid (Z,F) \in \mathcal{R^{+}} \right \} }(1-\sigma(Z))
    \times -\tau(G) \cdot \prod_{\left \{ Z \in \mathcal{A}\setminus{F} \mid (Z,G) \in \mathcal{R^{-}} \right \} }(1-\sigma(Z))\\
    &= \left[(1-0.5) \times (1-\left| 0-0 \right|) \right] \times (-0.5)\\
    &= -0.25.
\end{aligned}
$$
\end{example}

\section{Fraud Detection Example Details}

%\begin{example}[AAEs for a large QBAF]
\label{example_large_QBAF}
Consider the large QBAF from \cite{chi2021optimized} mentioned in the Introduction in Figure~\ref{fig_big_AF}, equipped with DF-QuAD gradual semantics. For ease of reference, in Figure~\ref{fig_QBAF_big_Example} below, we put indexes on the arguments. 
In this figure, we suppose argument $1$ (%It is a fraud case.
`We should start an investigation on this case as it is a fraud case' ) is the topic argument,  where argument $2$ (`It is %indeed 
a fraud case') supports and argument $3$ (%Actually, it 
`It is not a fraud case') attacks the topic argument%, respectively
.
According to the original paper~\cite{chi2021optimized}, the base score $\tau$ of every argument is set to $0.5$.
If $\sigma(1)>0.5$, we think the %case is fraudulent, else it is not
an investigation into the case should be started, else it does not need to.
Based on the DF-QuAD semantics, we have $\sigma(1)=0.2543945275247097<0.5$, which means %it is not a fraud case
the investigation does not need to start.
We next apply AAEs to quantitatively explain %the outcome of argument $1$ and 
this outcome and conduct some analysis. 

The corresponding argument contents and AAEs towards argument $1$ are shown in descending order in Table~\ref{tab:fraud_scores}.
From this table, we see that $29$ arguments have positive influence while $18$ arguments have negative influence on argument $1$. Among the former $29$ arguments, argument $2$ has the largest positive influence, while argument $3$ has the largest negative influence of all the latter $18$ arguments. 
Although there are more positive arguments than negative arguments, the average value of negative AAEs is larger than that of positive AAEs in terms of the absolute value, which roughly explains why $\sigma(1)<0.5$. 
In this case, using conversational explanations alone is not enough to explain $\sigma(1)$ because they fail to give quantitative reasons. Also, AAEs are %more intuitive 
leaner when explaining such large QBAF. Imagine using dialogues to explain the effect from argument $41$ to argument $1$: the explanations would be very lengthy% and lack of intuitiveness
. 
Besides, we find that some arguments share the same attribution score, like argument $6-11$: this is because these arguments are ``symmetric" in the QBAF, which faithfully reflects the real influence of them on the topic argument $1$. If we use qualitative conversational explanations alone, it would be difficult to compare the contributions between these arguments because conversational explanations are unable to measure and evaluate these contributions quantitatively.
\begin{figure}
	\centering
		\includegraphics[width=1.0\columnwidth]{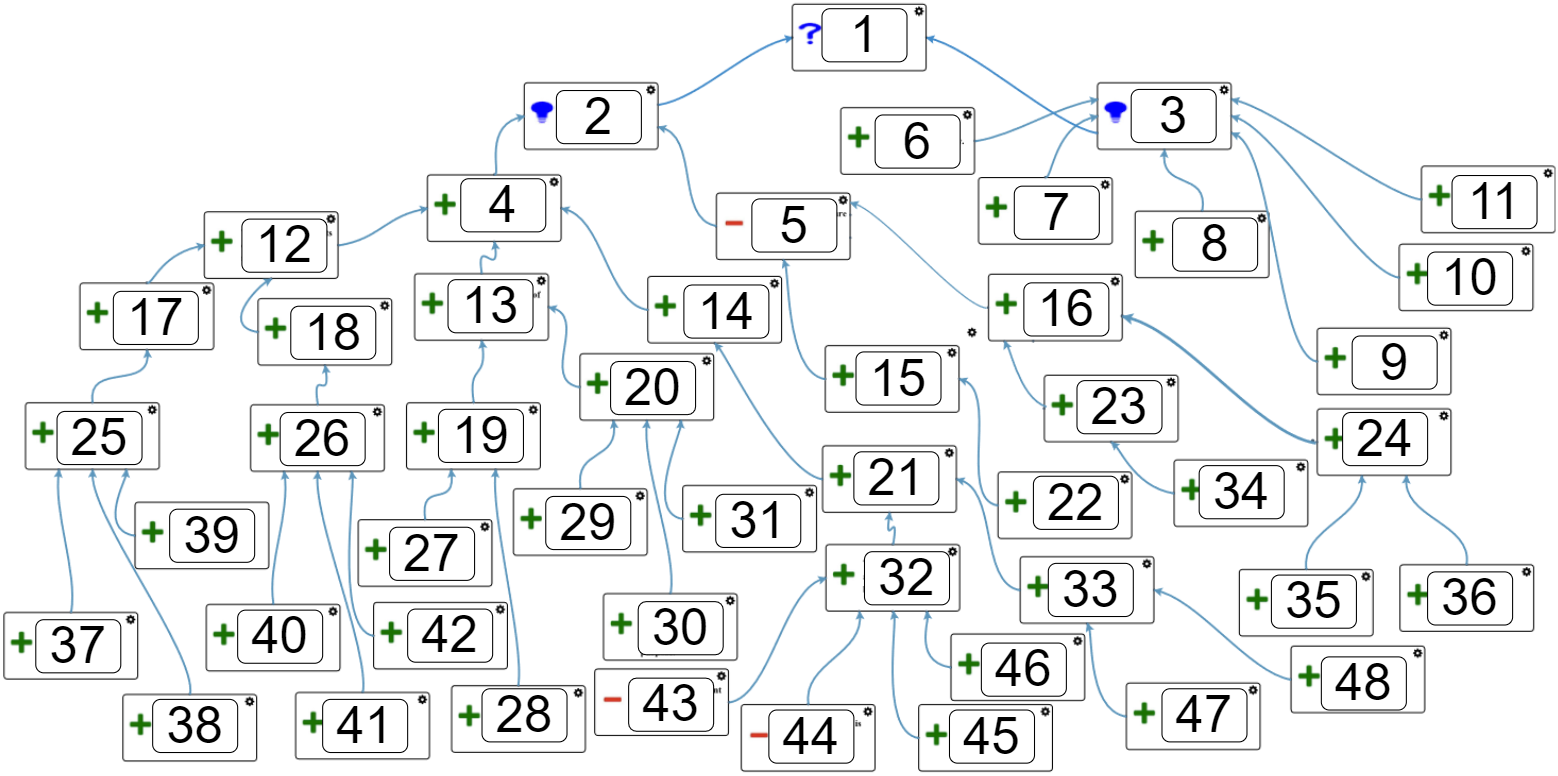}
	\caption{Fraud detection in e-commerce~\cite{chi2021optimized} with indexed arguments.}
	\label{fig_QBAF_big_Example}
\end{figure}

\begin{table}
\centering
\caption{Arguments, AAEs for topic argument 1 (`We should start an investigation on this case as it is a fraud case'), base scores and DF-QuAD strengths for the fraud detection QBAF.}
\label{tab:fraud_scores}
\begin{tabular}{ccccl}
\hline
\textbf{Index} & \textbf{AAE} & \textbf{Base score} & \textbf{Strength} & \textbf{Content} \\
\hline
2   & 4.99E-01 &0.5 &0.5009765550494194 & It is a fraud case. \\
4   & 1.46E-08 &0.5 &0.9999999850988388 & The fraud elements are satisfied. \\
27  & 3.66E-09 &0.5 &0.5 & The victim delivered the property voluntarily. \\
28  & 3.66E-09 &0.5 &0.5 & The victim delivered the property consciously. \\
45  & 3.66E-09 &0.5 &0.5 &  The collection behavior is suspicious. \\
46  & 3.66E-09 &0.5 &0.5 &  The collection account has high-risk record. \\
47  & 3.66E-09 &0.5 &0.5 &  Suspect removed victim from communication list. \\
48  & 3.66E-09 &0.5 &0.5 &  The suspect had no response. \\
29  & 1.83E-09 &0.5 &0.5 & The suspect refused to pay refund. \\
30  & 1.83E-09 &0.5 &0.5 & The suspect betrayed the victim's transaction purpose. \\
31  & 1.83E-09 &0.5 &0.5 & The property was transferred to the suspect. \\
37  & 1.83E-09 &0.5 &0.5 &  The suspect faked official materials. \\
38  & 1.83E-09 &0.5 &0.5 &  The transaction was at variance with business logic. \\
39  & 1.83E-09 &0.5 &0.5 &  The suspect faked official identity. \\
40  & 1.83E-09 &0.5 &0.5 &  The victim belongs to gullible group. \\ 
41  & 1.83E-09 &0.5 &0.5 & The transaction contained high-risk operations. \\ 
42  &1.83E-09	&0.5 &0.5 & The transaction was abnormal. \\ 
25	&9.15E-10	&0.5 &0.9375 &The basic facts of fabrication stand. \\ 
26	&9.15E-10	&0.5 &0.9375 &The victim had abnormal behaviours. \\ 
32	&9.15E-10	&0.5 &0.5 &The collection account is a blackaccount. \\ 
33	&9.15E-10	&0.5 &0.875 &The suspects terminated contact with the victim. \\ 
21	&4.58E-10	&0.5 &0.96875 &The suspect had illegal possession purpose. \\ 
19	&1.14E-10	&0.5 &0.875 &The victim's property has been delivered. \\ 
20	&1.14E-10	&0.5 &0.9375 &The suspect has obtained victim's property. \\ 
17	&1.43E-11	&0.5 &0.96875 &The suspect fabricates facts. \\ 
18	&1.43E-11	&0.5 &0.96875 &The victim fell into cognitive error. \\ 
12	&4.36E-16	&0.5 &0.99951171875 &The behavior elements of fraud cases are satisfied. \\ 
13	&4.36E-16	&0.5 &0.99609375 &The result elements of fraud cases are satisfied. \\ 
14	&4.36E-16	&0.5 &0.984375 &The suspect had direct intention for fraud. \\ 
43	&-3.66E-09	&0.5 &0.5 & The collection account is a mature account. \\ 
44	&-3.66E-09	&0.5 &0.5 & Collection account is a account for adaily use. \\ 
15	&-7.50E-06	&0.5 &0.75 &The auxiliary facts elements are satisfied. \\ 
16	&-7.50E-06	&0.5 &0.984375 &The victim has clear cognition. \\ 
23	&-6.00E-05	&0.5 &0.75 &The victim's complaints are malicious. \\ 
24	&-6.00E-05	&0.5 &0.875 &The victim had speculative motivitions. \\ 
6	&-2.40E-04	&0.5 &0.5 & The suspects is with limited capacity. \\ 
7	&-2.40E-04	&0.5 &0.5 & The suspect is with incapacity. \\ 
8	&-2.40E-04	&0.5 &0.5 & The suspect is under age of criminal responsibility. \\ 
9	&-2.40E-04	&0.5 &0.5 & The suspect has no capacity of criminal responsibility. \\  
10	&-2.40E-04	&0.5 &0.5 & The suspect had indirect intention. \\ 
11	&-2.40E-04	&0.5 &0.5 & The suspect fell into negligence. \\ 
35	&-4.80E-04	&0.5 &0.5 & The transaction is illegal. \\ 
36	&-4.80E-04	&0.5 &0.5 & The transaction occurred between acquitances. \\ 
22	&-9.60E-04	&0.5 &0.5 & The case is different from common fraud. \\ 
34	&-9.60E-04	&0.5 &0.5 & The victim's complaint is not credible. \\ 
5	&-1.92E-03	&0.5 &0.998046875 &The fraud elements are not satisfied. \\ 
3	&-7.81E-03	&0.5 &0.9921875 &It is not a fraud case. \\ 
\hline
\end{tabular}
\end{table}

%\end{example}

\end{document}